\documentclass[sn-mathphys-num]{sn-jnl}

\usepackage[table]{xcolor}
\usepackage{amssymb}
\usepackage{lipsum}
\usepackage{lscape}
\usepackage{graphicx}%
\usepackage{multirow}%
\usepackage{amsmath,amssymb,amsfonts}%
\usepackage{amsthm}%
\usepackage{amsfonts}
\usepackage{mathrsfs}%
\usepackage[title]{appendix}%
\usepackage{textcomp}%
\usepackage{manyfoot}%
\usepackage{booktabs}%
\usepackage{algorithm}%
\usepackage{algorithmicx}%
\usepackage{algpseudocode}%
\usepackage{listings}%
\usepackage{url}
\usepackage{physics}
\usepackage{amsmath}
\usepackage{tikz}
\usepackage{mathdots}
\usepackage{yhmath}
\usepackage{cancel}
\usepackage{color}
\usepackage{siunitx}
\usepackage{array}
\usepackage{multirow}
\usepackage{amssymb}
\usepackage{gensymb}
\usepackage{tabularx}
\usepackage{extarrows}
\usepackage{booktabs}
\usetikzlibrary{fadings}
\usetikzlibrary{patterns}
\usetikzlibrary{shadows.blur}
\usetikzlibrary{shapes}
\usepackage{listings}
\usepackage{graphicx} 
\usepackage{fancybox}
\usepackage{adjustbox}
\usepackage{twoopt} 
\usepackage{subcaption}
\usepackage{amsfonts} 
\usepackage{amssymb}
\usepackage{booktabs, multirow} 
\usepackage{soul}
\usepackage{changepage,threeparttable} 
\usepackage{forest}
\usepackage{pdflscape}
\usepackage{adjustbox}
\usetikzlibrary{shadows, trees, positioning}
\usepackage{booktabs}
\usepackage{xltabular}
\usepackage{tabularray}
\usepackage{enumerate}
\usetikzlibrary{shapes,arrows,positioning}
\usepackage[many]{tcolorbox}
\usetikzlibrary{calc}
\tcbuselibrary{skins}
\usepackage{dashrule}
\usetikzlibrary{decorations.pathreplacing}
\usepackage{titlesec}
\usepackage{float}
\usepackage{mathalpha}
\usepackage{dutchcal}
\usepackage{graphicx}
\usepackage{placeins}
\usepackage{pgfplots}
\pgfplotsset{compat=newest}

\renewcommand{\vec}[1]{\mathbf{#1}}
\newcommandtwoopt{\insertboxtwo}[3][1.0][4.0cm]{\noindent\fbox{\begin{minipage}[c]{#1\textwidth}\parbox[c][#2]{#1\textwidth}{#3}\hfill\end{minipage}}}

\newcommand{\accfun}[0]{\mathcal{m}}
\newcommand{\lossfun}[0]{\ell}
\newcommand{\epochset}[0]{\mathcal{E}}
\newcommand{\datasetset}[0]{\mathcal{D}}
\newcommand{\thetaset}[0]{\Theta}
\newcommand{\lossset}[0]{\mathcal{L}}
\newcommand{\accset}[0]{\mathcal{M}}
\newcommand{\evalfun}[0]{v}
\newcommand{\evalfunset}[0]{\mathcal{V}}

\newenvironment{insertedtext}{}{}%

\newcommand{\ucipairedplot}[5]{%
\begin{figure}[p]
    \centering

    \begin{subfigure}[t]{\textwidth}
        \centering
        \includegraphics[
            width=\textwidth,
            height=0.57\textheight,
            keepaspectratio
        ]{images/#1}
        \caption{Post-hoc checkpoint selection.}
        \label{#5-ph}
    \end{subfigure}

    \vspace{0.4em}

    \begin{subfigure}[t]{\textwidth}
        \centering
        \includegraphics[
            width=\textwidth,
            height=0.57\textheight,
            keepaspectratio
        ]{images/#2}
        \caption{Patience-based early stopping with $T=20$.}
        \label{#5-es}
    \end{subfigure}

    \caption{Relative checkpoint-selection gaps obtained on the UCI datasets using #3 as the training loss and #4 as the target metric. Columns correspond to the considered validation criteria, while colours identify the parameter-to-sample ratio $r$. Datasets are ordered according to their GDV.}

    \label{#5}
\end{figure}
}

\begin{document}

\title[Article Title]{\textit{Don't stop me now}: How Validation Criteria Affect Checkpoint Selection and Early Stopping}


\author[2]{\fnm{Andrea} \sur{Apicella}\footnotetext[0]{Corresponding author: Andrea Apicella, andapicella@unisa.it\\Preprint submitted to a journal for publication}}\email{andapicella@unisa.it}
\equalcont{These authors contributed equally to this work.}
\author[1]{\fnm{Francesco} \sur{Isgrò}}\email{francesco.isgro@unina.it}
\equalcont{These authors contributed equally to this work.}
\author[1]{\fnm{Andrea} \sur{Pollastro}}\email{andrea.pollastro@unina.it}
\equalcont{These authors contributed equally to this work.}
\author[1]{\fnm{Roberto} \sur{Prevete}}\email{roberto.prevete@unina.it}
\equalcont{These authors contributed equally to this work.}

\affil[1]{\orgdiv{Department of Electrical Engineering and Information Technology}, \orgname{University of Naples Federico II}, \orgaddress{\street{Via Claudio 21}, \city{Naples}, \postcode{80125}, \country{Italy}}}

\affil[2]{\orgdiv{Department of Information Engineering, Electrical Engineering, and Applied Mathematics (DIEM)}, \orgname{University of Salerno}, \orgaddress{\street{Via Giovanni Paolo II, 132}, \city{Fisciano (Salerno)}, \postcode{84084}, \country{Italy}}}




\abstract{Checkpoint selection is a standard component of neural network training, yet
the validation criterion used to select a checkpoint is often chosen
heuristically. Moreover, the same criterion may be used either only to rank
checkpoints after completion of a predefined training run or also to determine
when training should stop, thereby affecting both the selected checkpoint and
the set of checkpoints available for selection. In this work, we systematically
investigate the role of validation criteria under these two settings. We
separately vary the training loss, the validation criterion, and the target
evaluation metric, and compare post-hoc checkpoint selection, in which training
proceeds for all predefined epochs, with patience-based early stopping, in
which the validation criterion also controls training termination. We consider
three Cross-Entropy, C-Loss, and PolyLoss as training losses, and accuracy,
macro-F1, and Matthews correlation coefficient as target metrics. Experiments
are conducted on 26 UCI tabular datasets using controlled shallow neural
networks and on CIFAR-10 and TinyImageNet-200 using ResNet-50. Selection quality
is assessed through the relative gap between the test performance of the
validation-selected checkpoint and the best-observed test performance for the
same target metric over the complete predefined training run. Our results
show that no validation criterion is uniformly preferable across settings.
Under post-hoc selection, selecting checkpoints directly according to the
target validation metric often yields gaps comparable to or smaller than those
obtained using validation losses. Under patience-based early stopping, however,
the relative behaviour of the criteria changes substantially: metric-based
criteria can produce larger gaps, whereas alignment between the training and
validation losses is beneficial only under specific experimental conditions.
These findings show that the effectiveness of a validation criterion depends
not only on its relationship with the target metric, but also on whether it is
used solely to rank checkpoints or additionally to terminate training.}




\maketitle
\section{Introduction}

\begin{insertedtext}
In Artificial Neural Network (ANN) models, training and evaluation typically follow an iterative procedure in which each iteration corresponds to an \textit{epoch}, up to a predefined maximum number of epochs \citep{goodfellow2016deep}. The training procedure produces a sequence of parameter configurations, or \textit{checkpoints}, one for each training epoch. The model ultimately retained for deployment may either be the final checkpoint at the end of training or a checkpoint selected from the training trajectory using a checkpoint selection strategy applied on a held-out \textit{validation set}. Once selected, the checkpoint is evaluated on a fully unseen \textit{test set} using a target metric that defines the notion of performance relevant to the task at hand. In classification, for example, a model may be trained using the cross-entropy loss \cite{bishop2023deep} both for the training and the validation-based criterion while being evaluated using as target metric the accuracy, macro-F1, or Matthews correlation coefficient (MCC, \cite{chicco2023matthews}). 

This paradigm is standard in supervised  learning and underlies most modern experimental protocols
\citep{goodfellow2016deep}. The checkpoint selection process usually involves three quantities that are related but play distinct roles: the \textit{loss function} used to train the model; the \textit{validation criterion} which determines which checkpoint is considered preferable in terms of the model's generalization capability; and the \textit{target metric} used to evaluate the selected model.

\begin{insertedtext}
Notably, while the choice of training losses and target metrics is often made on the basis of well-known theoretical bases, by contrast the choice of the validation criterion used for checkpoint selection is not standardized and is often based on heuristic or ad hoc choices \cite{prechelt2002early,miseta2024surpassing}. 
For example, in classification, checkpoints can be assessed on the validation set using criteria that capture different aspects of predictive performance. 
\begin{insertedtext}  
Typically, it may be defined as the same loss used during training but evaluated on the validation set, as a different loss function, or directly in terms of the target evaluation metric.
\end{insertedtext}  
Moreover, the same validation criterion can be exploited not only to rank checkpoints but also to determine if training should stop before the predefined maximum number of epochs is reached. A common example is patience-based early stopping \citep{bishop1995regularization}: training is terminated when the selected validation criterion has not improved for a predefined number of consecutive epochs. The checkpoint attaining the best value of that criterion among those observed before termination is then retained. The validation criterion therefore plays two related but distinct roles under patience-based early stopping. First, its values are used to rank the observed checkpoints and select the one to retain. Second, its evolution over time is used by the stopping rule to determine the training termination. 
\begin{insertedtext}
Prechelt \cite{prechelt1998automatic,prechelt2002early} explicitly recognizes the distinction between the epoch at which training is terminated and the checkpoint retained at the end of training. However, across all the considered stopping criteria, the retained checkpoint is the one attaining the the best validation criterion value observed before termination. Building on these observations, we conduct an experimental study by varying the validation function used for checkpoint selection and by examining how its role changes when it is used only to rank checkpoints or also to determine when training terminates.
\end{insertedtext}
Consequently, changing the validation criterion can affect both the selected checkpoint and the portion of the training trajectory that is explored. Therefore, different criteria may lead to different preferred checkpoints, and there is no universally established prescription as to whether checkpoint selection should be based on the training loss evaluated on the validation set, on a different loss, or directly on the target evaluation metric, or other criteria.
Furthermore, the possibility that these criteria select different checkpoints does not, by itself, reveal how much the final model performance is lost as a consequence of the validation-based decision, how this loss varies across experimental conditions, or whether the relative ranking of the criteria changes when they are also used to terminate training.
All these aspects constitute the central focus of this study.

\end{insertedtext}



\end{insertedtext}

\begin{insertedtext}
In particular, we conduct a controlled empirical study in which the loss used during the training to generate the checkpoints, the validation-based criterion used to evaluate checkpoints, and the metric used to assess the final selected checkpoint are varied independently. We compare two different validation-based protocols: (i) in \textit{post-hoc checkpoint selection}, training proceeds for the predefined maximum number of epochs and the validation criterion selects a checkpoint from the complete trajectory, while (ii) in \textit{patience-based early stopping}, the same criterion is used both to determine when training is terminated and to retain the best checkpoint observed before termination. This design allows us to assess whether conclusions obtained when all checkpoints are available remain valid when the validation signal also controls the duration of training.

Models are trained for classification tasks, using three alternative common loss functions proposed in literature, that are cross-entropy \cite{bishop2023deep}, C-Loss \citep{singh2010loss,singh2014c}, and PolyLoss \cite{leng2022polyloss}. The selected checkpoints are evaluated using accuracy, macro-F1 \cite{hastie2009elements}, and MCC as alternative target metrics \cite{chicco2023matthews}. Checkpoints are then ranked on the validation set using each of these losses or the target metric. We quantify the performance gap between the checkpoint selected using validation data and the best test performance observed retrospectively across all epochs up to the predefined maximum number of training epochs, examining how this gap varies across training losses, validation criteria, target metrics, datasets, and selection protocols.
The experiments cover both tabular and image-classification tasks. For the tabular tasks, we use benchmark classification datasets from the UCI Machine Learning Repository~\citep{asuncion2007uci}. Motivated by the analysis of the relationship between the number of model parameters and generalization provided by \citet{advani2020high}, we employ fully connected neural networks with a single hidden layer and vary the number of model parameters with respect to the dataset size. Instead, for image-based classification tasks, we used CIFAR-10 and TinyImageNet-200 with a ResNet50 architecture \cite{he2016deep}.
\end{insertedtext}

\begin{insertedtext}
The results do not support a universally optimal validation criterion. Under post-hoc checkpoint selection, selecting checkpoints directly according to the target validation metric frequently produces performance gaps comparable to, or smaller than, those obtained using validation losses. Instead, under patience-based early stopping, the ranking of the criteria changes substantially in several settings. In particular, metric-based evaluation can produce larger gaps, whereas matching the training and validation losses was beneficial in some cases, but this advantage does not hold consistently across training objectives, target metrics, and datasets. These findings indicate that the effectiveness of a validation criterion depends not only on its alignment with the target evaluation metric, but also on whether it is used exclusively to rank checkpoints or additionally to terminate training.

Accordingly, the main contributions of this work are:

\begin{itemize}
    \item an experimental framework that separately accounts for the training loss, the validation criterion, and the target evaluation metric, and compares the use of the validation criterion for checkpoint selection alone with its joint use for checkpoint selection and training termination;
    \item a quantitative characterization of the performance gap between a validation-selected checkpoint and the best test performance observed retrospectively along the corresponding training trajectory;
    \item a comparison between post-hoc checkpoint selection and patience-based early stopping, showing that the relative ranking of validation criteria does not necessarily transfer from one protocol to the other;
    \item an empirical evaluation across multiple training losses, validation criteria, target metrics, number of model parameters, and both tabular and image-classification datasets.
\end{itemize}
\end{insertedtext}

The remainder of the paper is organized as follows. Section~\ref{sec:related} reviews related works; Section~\ref{sec:method} introduces notation, the evaluation criteria and experimental protocol; Section~\ref{sec:exp} presents datasets, models, and evaluation procedures; Section~\ref{sec:results} reports and discusses the results and their limitations; and Section~\ref{sec:conclusions} concludes the work with final remarks.

\section{Related Work}
\label{sec:related}
In supervised learning, models are trained on labeled data belonging to a given task, with the aim to achieve high values of a task performance measure \citep{terven2025comprehensive}. However, in several tasks the target metric is often non-differentiable (e.g., accuracy, F1) or yields flat/unstable gradients for gradient-based optimization. Consequently, training relies on differentiable \emph{surrogate losses} (e.g., cross-entropy) that act as proxies for the task metric. However, minimizing the training loss does not guarantee improvements in the deployment evaluation metric (loss–metric mismatch, \citep{huang2019addressing}). Motivated by this gap, several works design losses that more closely reflect the task objectives: for example, the C-Loss based on cross-correntropy as a surrogate to the 0-1 risk \citep{santamaria2006generalized,singh2010loss}, or probabilistic performance indices that jointly reward correctness, high probability for the true class, and low probability for the others \citep{wang2013probabilistic}, or score-oriented losses that target confusion-matrix summaries \citep{marchetti2022score,marchetti2025comprehensive}. 
\cite{ru2020revisiting} propose accumulated training loss as an efficient estimator for ranking architectures in Neural Architecture Search (NAS). Their study compares this estimator with early validation accuracy and validation-loss-based alternatives. However, their objective is to predict the final performance ranking across different architectures from partially completed training runs. In contrast, our study examines the selection of a checkpoint within a given training run and quantifies the resulting target-metric gap under both post-hoc checkpoint selection and patience-based early stopping.
Furthermore, \cite{venter2020pre} showed that validation loss can increase while validation accuracy remains stable or even improves, due to a small subset of samples exhibiting extremely large loss values. Their analysis suggests that averaged validation loss and classification accuracy may encode different information about model behavior and generalization dynamics.

Other works introduced task- and data-adaptive loss functions (e.g., PolyLoss \citep{leng2022polyloss}), defining parametric families in which standard objectives, such as cross-entropy, arise as special cases.
Regardless of the training objective, generalization is ultimately assessed on held-out data (validation/test). While the test set is usually evaluated using the effective task-specific metric,  the criterion used on the validation set can suffer from the same loss-metric mismatch: selecting \textit{checkpoints}, i.e., the epoch corresponding to the model parameters to be selected, by a surrogate such as cross-entropy may fail to identify the model that maximizes the task-specific metric (e.g., accuracy). PACMAN~\citep{vera2024pacman} provides generalization bounds that explicitly account for the discrepancy between cross-entropy and accuracy, while other works address the loss-metric mismatch through adaptive loss design~\citep{huang2019addressing} or empirical analyses of generalization behavior~\citep{liao2018surprising}. However, these approaches do not directly examine the implications of this mismatch for validation-based model selection. In contrast, our work focuses on the effectiveness of model selection procedures driven by validation criteria, explicitly comparing validation-selected models against the test-optimal model under controlled experimental settings.

Furthermore, when models are evaluated iteratively on a validation set, it is common to use early stopping to truncate optimization before convergence \citep{bishop1995regularization}. It was shown that stopping early can yield solutions comparable to those of smaller, optimally sized models \citep{caruana2000overfitting}, and consistency results are available under specific assumptions \citep{ji2021early}. The benefit of early stopping depends on the loss and the geometry of the optimization landscape: studies of loss surfaces and representation dynamics highlight plateaus, saddle points, overconfidence, and how regularization shapes hidden-layer encodings \citep{soudry2016no,swirszcz2016local,goodfellow2014qualitatively,zhang2020penetrating}. 
Classical analyses investigated overtraining dynamics for linear networks under quadratic loss and characterized validation-based stopping both geometrically and in time \citep{baldi1991temporal,wang1993optimal,dodier1995geometry}, while statistical views related early stopping to explicit regularization \citep{hagiwara2002regularization,evgeniou2002regularization}.
Berta et al.~\citep{berta2025rethinking} discuss how validation loss itself may represent a suboptimal compromise between refinement and calibration errors during training, proposing a refined stopping criterion.
In contrast to refinement-based approaches, we focus on the empirical discrepancy between validation-selected checkpoints and the test-optimal checkpoint under different validation criteria.

Guan et al. \cite{guan2025systematic} conduct a systematic study of early stopping metrics in the context of hyperparameter optimization (HPO). While their analysis is framed within Hyperparameter Optimization (HPO) and resource-allocation strategies such as Hyperband and Successive Halving, our work focuses on a complementary problem, i.e. the reliability of validation criteria for checkpoint selection and early stopping in standard supervised training pipelines. However, their work similarly highlights that the choice of early stopping metric has significant and non-obvious consequences for model selection outcomes, and that commonly assumed defaults could not always be the optimal strategies.

It is interesting to notice that optimal–stopping effects appear beyond artificial neural networks, notably in boosting methods \citep{buhlmann2003boosting,barron2008approximation,chen2013learning,wei2017early}, and SVMs \citep{bandos2007statistical}, which helps explain the widespread use of early stopping. 
More in general, stopping rules can be applied either on training data or on a held-out validation set \citep{ferro2023early}. Training-monitored criteria include, for example, a log-sensitivity index for rare outcomes \citep{ennett2003evaluation} and rules driven by training-loss trajectories \citep{lalis2014adaptive}; protocol-centric choices around how the hold-out split is constructed have also been explored \citep{wu2009new}. However, some early procedures relied solely on training-set criteria or repeatedly re-sampled ``validation'' from the training pool \citep{natarajan1997automated,iyer2000novel}, practices that can bias selection and inflate performance estimates \citep{apicella2025don}.  Validation-monitored rules span comparative studies and benchmarks of families and combinations \citep{prechelt1998automatic,prechelt2002early,lodwich2009evaluation,nguyen2005stopping}, as well as practical heuristics such as fixed validation-error thresholds \citep{suliman2018early} or marginal-improvement criteria \citep{shao2010comparison}.
\begin{insertedtext}
 A particularly relevant empirical study is provided by \citet{prechelt1998automatic, prechelt2002early}, who compares 14 validation-based stopping criteria across multiple tasks. Our work addresses a complementary question. Rather than comparing alternative stopping rules defined from the evolution of validation error and training progress, we examine how different loss-based and target-metric-based validation functions affect the retained checkpoint. We additionally distinguish their behaviour when they are used only for post-hoc checkpoint ranking from their behaviour when they also control training termination, and quantify the resulting target-metric gap relative to the best-observed test performance.
\end{insertedtext} 

In summary, prior work highlights three themes that motivate our study: (i) models are trained with surrogate losses that may not align with task metrics; (ii) validation criteria inherit this mismatch and thus critically determine which checkpoint is selected; and (iii) early stopping is often used to avoid running all epochs, so the validation signal effectively chooses the model instance among the per-epoch checkpoints—making the choice of validation metric especially important. Our study addresses these themes within a unified experimental protocol by comparing loss-based checkpoint selection, using Cross-Entropy, C-Loss, or PolyLoss, with selection based directly on the corresponding target validation metric, namely accuracy, macro-F1, or MCC.

\section{Method}
\label{sec:method}
\subsection{Notation}
\label{subsec:notation}

In this work, we adopt the following notation. In supervised ML, considering a classification problem, a dataset $ D \in \mathcal{D} $ consists of $ N $ input-label pairs $D = \{ (\mathbf{x}^{(i)}, y^{(i)}) \}_{i=1}^N,$ where $ \mathbf{x}^{(i)} $ denotes an input instance and $ y^{(i)} $ the corresponding ground-truth value. 
The set $\datasetset $ denotes the collection of all possible datasets for the task under consideration.


We focus on a classification setting in which each input  $\mathbf{x}^{(i)}$ is assigned to one of $K$ mutually exclusive classes  $\{1, 2, \dots, K\}$. For simplicity and without loss of generality, labels are treated as one-dimensional discrete values, i.e., $y^{(i)} \in \{1, 2, \dots, K\}.$ 
\begin{insertedtext}
    
Given an ML model with trainable parameters $\theta$, let $\epochset=\{1,\ldots,E\}$ denote the set of training epochs. A training run produces a sequence of parameter configurations $\thetaset=\{\theta_e\}_{e\in\epochset}$, where $\theta_e$ denotes the parameters obtained at the end of epoch $e$. 

Let $\lossfun :\datasetset \times \thetaset \to \mathbb{R}\quad \text{and} \quad\accfun : \datasetset \times \thetaset \to  \mathbb{R}$ denote respectively a loss function and a task-related evaluation metric function computed on a dataset $D\in \datasetset$ on the model with parameter values  $\theta\in\thetaset$. With $\lossset$ and $\accset$, we denote, respectively, the sets of loss functions and target evaluation metrics considered in this study.
For notational simplicity and without loss of generality, we assume that all these functions are represented so that larger values are preferable. We then define $\evalfunset=\lossset\cup\accset$ as the set of evaluation functions considered for checkpoint selection. 
\end{insertedtext}

\subsection{Validation-based Checkpoint Selection Protocol}
\begin{insertedtext}
In this work, we distinguish between two validation-based checkpoint selection protocols: (i) Training proceeds for a fixed number of epochs. Once training is complete, a given checkpoint-evaluation function is used on the validation set to select the checkpoint with the best validation score among all those generated during training. Here we refer to this as \textit{post-hoc checkpoint selection}; (ii) During training, a stopping criterion is monitored, and training is halted when that criterion is satisfied. A very common stopping criterion is \textit{patience-based early stopping}, where the checkpoint achieving the best score according to a given checkpoint-evaluation function is selected from among those generated up to the stopping epoch. Thus, in the following paragraphs we formalize these two validation-based checkpoint selection protocols.
\paragraph{\emph{Post-hoc Checkpoint Selection (PH)}} 
For a dataset $D\in \datasetset$ and a checkpoint-evaluation function $v$, the training proceeds for all the epochs in $\epochset=\{1,2,\dots,E\}$ and the checkpoint $\theta_{e^{\star}_{v, D}}$ is selected retrospectively as the one at iteration $e^{\star}_{v, D}=\arg\max\limits_{e\in\epochset}v(D, \theta_e)$. 

Accordingly, we define the best-observed evaluation value along the training trajectory $\epochset$ as
$v_D^\star= v(D,\theta_{e^{\star}_{v, D}})$.
Here, the superscript $\star$ denotes the best value observed among the available checkpoints and should not be interpreted as a theoretical or global optimum.

\begin{figure}
    \centering
    \begin{tikzpicture}

\definecolor{darkgray176}{RGB}{176,176,176}
\definecolor{darkorange25512714}{RGB}{255,127,14}
\definecolor{lightgray204}{RGB}{204,204,204}
\definecolor{steelblue31119180}{RGB}{31,119,180}

\begin{axis}[
legend cell align={left},
legend style={
    fill opacity=0.8,
    draw opacity=1,
    text opacity=1,
    at={(0.03,0.97)},
    anchor=north west,
    draw=lightgray204,
    font=\scriptsize
},
tick align=outside,
tick pos=left,
x grid style={darkgray176},
xlabel={$\scriptsize \text{epoch } e$},
xmajorgrids,
xmin=1,
xmax=80,
xtick style={color=black},
xtick={38,60,80},
xticklabels={
    $e^\star_{v_\ell,Val}$,
    $e^\star_{v_m,Val}$,
    $E$
},
xticklabel style={font=\scriptsize},
y grid style={darkgray176},
ylabel={Raw criterion value},
ymajorgrids,
ymin=0.220747551135609,
ymax=2.12560040789433,
ytick style={color=black}
]

\addplot [semithick, steelblue31119180]
table {%
1 1.48851100210574
2 1.44883574733409
3 1.36150722276629
4 1.33756242805696
5 1.25724597640264
6 1.18361452003697
7 1.11645678841415
8 1.0481238907509
9 0.995385402946758
10 0.950922201876592
11 0.900830158224857
12 0.861216377915655
13 0.804006976335615
14 0.775567182830839
15 0.778761754117512
16 0.743930709084563
17 0.72414520162041
18 0.678461351066591
19 0.616238417591064
20 0.596765870239478
21 0.547093540747008
22 0.501962197622451
23 0.455337468325465
24 0.434067360344413
25 0.410964257608586
26 0.407585880424113
27 0.392749036143047
28 0.404687925521248
29 0.380315493223299
30 0.391301266213936
31 0.387393961431952
32 0.365790072386484
33 0.3509530309652
34 0.358774257360046
35 0.339358384381129
36 0.32037775312598
37 0.318585997467318
38 0.307331771897369
39 0.323733298542357
40 0.330230805686256
41 0.359468365944972
42 0.392910137558864
43 0.4066329372407
44 0.444418880526643
45 0.44835458720942
46 0.474410781981422
47 0.478184546629683
48 0.502557914502292
49 0.515424681678488
50 0.525685495274196
51 0.542483021671747
52 0.582505008974042
53 0.614715936527399
54 0.626948399079629
55 0.695408069196528
56 0.714741977628116
57 0.774045157306287
58 0.827636347055936
59 0.875070060763644
60 0.903709089253858
61 0.943023940524156
62 0.98426947871063
63 1.00923564251718
64 1.0444988771952
65 1.09696273897276
66 1.14563422555802
67 1.18935743389933
68 1.25122580942893
69 1.31806570007525
70 1.38488781511354
71 1.45683833457579
72 1.52293747641665
73 1.60810328581716
74 1.66511155913922
75 1.7252053452386
76 1.78689700816779
77 1.8484443235595
78 1.91818992078111
79 1.96723063783179
80 2.03901618713257
};
\addlegendentry{
    Validation loss $\lossfun(Val,\theta_e)$
    ($\downarrow$)
}

\addplot [semithick, darkorange25512714]
table {%
1 0.677160627422549
2 0.686325173616671
3 0.706497099470024
4 0.71202808291278
5 0.730580297273757
6 0.747588351465971
7 0.763101046655875
8 0.778885192241484
9 0.791067201180063
10 0.80133771018661
11 0.812908419779201
12 0.82205876621601
13 0.835273507262976
14 0.841842786611801
15 0.841104877502296
16 0.849150468035475
17 0.853720708838562
18 0.864273188083643
19 0.878646026388824
20 0.883144022573052
21 0.894617883181201
22 0.905042914469917
23 0.915813062108157
24 0.920726858715086
25 0.926064560392798
26 0.926846937007537
27 0.930277567257479
28 0.927525756365091
29 0.933165488034633
30 0.930644350074379
31 0.931573609909957
32 0.936606533487836
33 0.940100735603291
34 0.938397625204504
35 0.943039704025791
36 0.947658859544135
37 0.948417611885027
38 0.951515135488852
39 0.948433227578071
40 0.947918117649323
41 0.942515858071051
42 0.936611300220489
43 0.935849912922584
44 0.93025185386839
45 0.933337187030739
46 0.932322391978723
47 0.937601950362935
48 0.939389239378671
49 0.945185866681643
50 0.952973065607469
51 0.960613148729729
52 0.964149643176669
53 0.970564321082779
54 0.982391283022385
55 0.981664788140189
56 0.992280886860584
57 0.993162276688665
58 0.994330895103412
59 0.995358098515919
60 0.998663916807094
61 0.997008768269562
62 0.992078444685237
63 0.987867948595894
64 0.978166118026437
65 0.961450603928459
66 0.942781878999586
67 0.922761182228934
68 0.896486401915121
69 0.867499532405434
70 0.837484590787319
71 0.805782796182943
72 0.775427689822584
73 0.741102801463271
74 0.714079472332895
75 0.68741733860833
76 0.661647118566675
77 0.637273033960991
78 0.612393861779998
79 0.593648728978534
80 0.57091580256769
};
\addlegendentry{
    Validation metric $\accfun(Val,\theta_e)$
    ($\uparrow$))
}

\addplot [
    line width=0.48pt,
    steelblue31119180,
    opacity=0.85,
    dash pattern=on 4.44pt off 1.92pt,
    forget plot
]
table {%
38 0.220747551135609
38 2.12560040789433
};

\addplot [
    line width=0.48pt,
    darkorange25512714,
    opacity=0.85,
    dash pattern=on 4.44pt off 1.92pt,
    forget plot
]
table {%
60 0.220747551135609
60 2.12560040789433
};

\addplot [
    line width=0.48pt,
    steelblue31119180,
    opacity=0.85,
    dash pattern=on 1.2pt off 1.98pt,
    forget plot
]
table {%
1 0.307331771897369
80 0.307331771897369
};

\addplot [
    line width=0.48pt,
    darkorange25512714,
    opacity=0.85,
    dash pattern=on 1.2pt off 1.98pt,
    forget plot
]
table {%
1 0.998663916807094
80 0.998663916807094
};

\draw[->,draw=black]
(axis cs:33,0.427331771897369) --
(axis cs:38,0.307331771897369);

\draw (axis cs:33,0.427331771897369) node[
    scale=0.5,
    anchor=south east,
    text=black
]{
    $(e^\star_{v_\ell,Val},L^\star_{Val})$
};

\draw[->,draw=black]
(axis cs:65,1.30) --
(axis cs:60,0.998663916807094);

\draw (axis cs:65,1.30) node[
    scale=0.5,
    anchor=south west,
    text=black
]{
    $(e^\star_{v_m,Val},M^\star_{Val})$
};

\end{axis}
\end{tikzpicture}
    \caption{Illustrative validation-loss and validation-metric trajectories computed along the same training trajectory. In this figure, we adopt the conventional orientation in which the loss function $\lossfun(Val,\theta_e)$ has to be  minimized ($\downarrow$), whereas the evaluation metric $\accfun(Val,\theta_e)$ has to be maximized ($\uparrow$).  Accordingly,$e^\star_{\lossfun,Val}$ denotes the epoch minimizing the validation loss, while $e^\star_{\accfun,Val}$ denotes the epoch maximizing the validation metric. The vertical dashed lines indicate the corresponding selected epochs, whereas the horizontal dotted lines indicate the best-observed validation values $\lossfun^\star_{Val}$ and $\accfun^\star_{Val}$. Displaying the loss in its original orientation is equivalent to the common maximization convention adopted in the formal analysis, since maximizing $-\lossfun$ selects the same checkpoint as minimizing $\lossfun$.}
    \label{fig:ex1}
\end{figure}

\paragraph{\emph{Patience-based Early Stopping (ES)}}
With early stopping, the training is terminated according to a predefined stopping criterion. A commonly adopted strategy is \emph{early stopping with patience $T$}, whereby training is halted when no strict improvement in the checkpoint evaluation function $v\in\evalfunset$ has been observed for $T$ consecutive epochs.

More formally, for a dataset $D\in \datasetset$ and  a chosen checkpoint evaluation function $v\in \evalfunset$, given the set of training epochs $\epochset=\{1,2,\dots,E\}$ and a patience parameter $T\in\mathbb{N}$, with $T \leq E$, we can define an early stopping predicate $P_{\mathrm{stop}}$ as
$$P_{\mathrm{stop}}(t)=\begin{cases}\mathrm{true}, & (t=E)  \lor \left(\big(t > T\big)\land  \big(v(D,\theta_{t-T})\geq\max\limits_{t-T+1\leq j\leq t}v(D,\theta_{j})\big)\right),\\ \mathrm{false}, & \text{otherwise.}
\end{cases}
$$
Thus, excluding the maximum-epoch condition $t=E$, $P_\mathrm{stop}(t)$ becomes true when the  checkpoint evaluation function has not improved for $T$ consecutive epochs on $D$. In particular, equality is not regarded as an improvement. The stopping epoch is defined as the first epoch at which the predicate becomes true: $$t_{\mathrm{stop}}=\min\left\{t\in\epochset:P_\mathrm{stop}(t)=\mathrm{true}\right\}.$$ 
The epoch of the selected checkpoint is usually defined as 
$$\hat{t}=\arg\max_{1\leq j\leq t_{\mathrm{stop}}}
v(D,\theta_j)$$
When $t_{\mathrm{stop}}<E$, the first occurrence of the early-stopping condition guarantees that $v(D,\theta_{t_{\mathrm{stop}}-T})$ is a maximum over all epochs completed up to $t_{\mathrm{stop}}$. When $t_{\mathrm{stop}}=E$, the best checkpoint is necessarily contained among the last $T+1$ completed epochs. If the maximum is attained at multiple epochs, a fixed tie-breaking rule is used.
The retained checkpoint is $\theta_{\hat{t}}$, which achieves the best validation value observed up to the stopping epoch $t_{\mathrm{stop}}$.
\end{insertedtext}

Figure~\ref{fig:ex2} depicts both procedures, i.e., post-hoc checkpoint selection and early stopping.

Note that $T=E$ represents a degenerate limiting case, i.e., when the patience parameter equals the total number of epochs, patience-based early stopping results in no early termination and is therefore equivalent to post-hoc checkpoint selection.

 \begin{figure}
    \centering
    \begin{tikzpicture}

\definecolor{darkgray153}{RGB}{153,153,153}
\definecolor{darkgray176}{RGB}{176,176,176}
\definecolor{darkorange25512714}{RGB}{255,127,14}
\definecolor{lightgray204}{RGB}{204,204,204}
\definecolor{steelblue31119180}{RGB}{31,119,180}

\begin{axis}[
legend cell align={left},
legend style={
    fill opacity=0.8,
    draw opacity=1,
    text opacity=1,
    draw=lightgray204,
    at={(0.50,0.18)},
    anchor=north west,
    font=\scriptsize
},
tick align=outside,
tick pos=left,
x grid style={darkgray176},
xlabel={$\text{epoch }t$},
xmajorgrids,
xmin=1,
xmax=60,
xtick style={color=black},
xtick={18,26,48,60},
xticklabels={
    $\hat{t}_{\mathrm{ES}}(\lossfun)$,
    $t_{\mathrm{stop}}$,
    $\hat{t}_{\mathrm{PH}}(\lossfun)$,
    $E$
},
xticklabel style={
    font=\scriptsize,
    align=center
},
y grid style={darkgray176},
ylabel={$\lossfun(Val,\theta_t)$},
ymajorgrids,
ymin=0.26360723624675,
ymax=0.909958525469701,
ytick style={color=black}
]

\addplot [
    semithick,
    steelblue31119180
]
table {%
1 0.86482449320904
2 0.873101656860381
3 0.885060555931067
4 0.877187083356736
5 0.876475359529115
6 0.865599514009249
7 0.846200934082393
8 0.838684568449853
9 0.81759025820095
10 0.801519250633917
11 0.789167852583789
12 0.766835356933559
13 0.773646305665266
14 0.756503333341052
15 0.764105551348646
16 0.759603833002953
17 0.770959732567684
18 0.662002270721324
19 0.722002270721306
20 0.73200227072117
21 0.742002270720219
22 0.75200227071404
23 0.762002270676654
24 0.772002270466245
25 0.782002269364784
26 0.792002264002328
};

\addlegendentry{
    before stopping
}

\addplot [
    semithick,
    steelblue31119180,
    opacity=0.45,
    dashed
]
table {%
26 0.792002264002328
27 0.721305472184363
28 0.704852891938667
29 0.687068346329864
30 0.673638978313501
31 0.663711225431872
32 0.654543099415848
33 0.669034432689344
34 0.651586098037879
35 0.667570154581956
36 0.676836857697005
37 0.686283057705817
38 0.694244200044216
39 0.687357793932619
40 0.67045209644255
41 0.658696949899141
42 0.619292980570674
43 0.598913566795935
44 0.546276290571824
45 0.497968381976598
46 0.456856224636922
47 0.429087768415854
48 0.410813517099933
49 0.427754900469004
50 0.426613034777813
51 0.460196299785715
52 0.495930183874823
53 0.53276767477077
54 0.566299203490648
55 0.590071832636656
56 0.608766966504567
57 0.618589191541326
58 0.619099813442012
59 0.627292823282023
60 0.624651096380606
};

\addlegendentry{
    continuation
}

\addplot [
    line width=0.6pt,
    darkorange25512714,
    dash pattern=on 4.44pt off 1.92pt,
    forget plot
]
coordinates {
    (18,0.26360723624675)
    (18,0.909958525469701)
};

\addplot [
    line width=0.6pt,
    darkgray153,
    dash pattern=on 2pt off 1.5pt,
    forget plot
]
coordinates {
    (26,0.26360723624675)
    (26,0.909958525469701)
};

\addplot [
    line width=0.6pt,
    steelblue31119180,
    dash pattern=on 4.44pt off 1.92pt,
    forget plot
]
coordinates {
    (48,0.26360723624675)
    (48,0.909958525469701)
};

\addplot [
    only marks,
    mark=*,
    mark size=2pt,
    darkorange25512714,
    forget plot
]
coordinates {
    (18,0.662002270721324)
};

\addplot [
    only marks,
    mark=*,
    mark size=2pt,
    steelblue31119180,
    forget plot
]
coordinates {
    (48,0.410813517099933)
};

\draw (axis cs:8,0.632002270721324) node[
    scale=0.55,
    fill=white,
    draw=darkorange25512714,
    line width=0.4pt,
    inner sep=2pt,
    fill opacity=0.9,
    text opacity=1,
    rounded corners,
    anchor=south west,
    text=black
]{
    early-stopping checkpoint
};

\draw (axis cs:26,0.835) node[
    scale=0.55,
    fill=white,
    draw=darkgray153,
    line width=0.4pt,
    inner sep=2pt,
    fill opacity=0.9,
    text opacity=1,
    rounded corners,
    anchor=west,
    text=black
]{
    training stops
};

\draw (axis cs:47,0.410813517099933) node[
    scale=0.55,
    fill=white,
    draw=steelblue31119180,
    line width=0.4pt,
    inner sep=2pt,
    fill opacity=0.9,
    text opacity=1,
    rounded corners,
    anchor=north east,
    text=black
]{
    post-hoc checkpoint
};

\draw[<->,draw=black]
(axis cs:18,0.298115822027297) --
(axis cs:26,0.298115822027297);

\draw (axis cs:22,0.287160715430298) node[
    scale=0.55,
    anchor=north,
    text=black
]{
    $T$ epochs
};

\end{axis}
\end{tikzpicture}
    \caption{Illustration of patience-based early stopping and post-hoc checkpoint selection on the same training trajectory. The figure displays the validation loss in its classical minimization form. In this example, the best validation loss observed up to $\hat{t}_{PS}$ is not improved during the following $T$ epochs. Training therefore stops at $t_{\mathrm{stop}}$, and early stopping retains checkpoint at epoch $\hat{t}$. Under fixed-horizon training, the trajectory continues until $E$, and post-hoc selection retains checkpoint $\theta_{PH}$. The dashed portion of the trajectory would not be observed under the actual early-stopping protocol and is shown only to illustrate the difference between the two protocols.}
    \label{fig:ex2}
\end{figure}



\subsection{Comparison of Model Selection Criteria}
\begin{insertedtext}
To assess the effect of validation criteria on generalization performance, we systematically compare loss-based and target-metric-based checkpoint selection under both post-hoc selection and patience-based early stopping. 

Our goal is to quantify the target-metric performance gap associated with validation-based checkpoint selection. Model parameters are optimized using the training set $Train\in\datasetset$, while the validation set $Val\in\datasetset$ is used exclusively to select the checkpoint. The test set $Test\in\datasetset$ is then used to evaluate the generalization performance of the selected checkpoint. More specifically, the gap compares the test performance of the validation-selected checkpoint with the best-observed value of the same target metric along the complete training trajectory. Let $q\in\mathcal{Q}=\{\mathrm{PH},\mathrm{ES}\}$ denote the adopted validation-driven protocol, where $\mathrm{PH}$ and $\mathrm{ES}$ indicate post-hoc checkpoint selection and early stopping, respectively. For a validation criterion $v\in\evalfunset$, let $\hat e_{v,Val}^{\,q}$ denote the epoch selected under protocol $q$ on the validation set $Val$. For a target metric $m\in\mathcal{M}$, the test performance of the selected checkpoint is therefore
$m\!\left(Test,\theta_{\hat e_{v,Val}^{\,q}}\right).$

We compare this value with the best-observed value of the same target metric along the complete training trajectory,
$m_{Test}^{\star}=\max_{e\in\epochset}m(Test,\theta_e)$.
Thus, the comparison measures how much target-metric performance is lost by selecting a checkpoint according to $v$ on the validation set. The quantity $m_{Test}^{\star}$ is used exclusively as a retrospective reference for quantifying this gap; test-set information is never used for checkpoint selection.
\end{insertedtext}

\begin{insertedtext}
More in detail, for each dataset, data split, and training objective $\lossfun_{\mathrm{train}}\in\lossset$, the model was trained for a maximum of $E$ epochs, and the corresponding sequence of checkpoints $(\theta_e)_{e\in\epochset}$ was recorded. At every epoch, the validation values of the considered loss functions and target evaluation metrics were computed.

Since our aim is to compare loss-based checkpoint selection with checkpoint selection based directly on the same task-related evaluation metric,
under the common maximization convention introduced above, for each target metric $\accfun\in\accset$, we restrict the set of candidate validation criteria to
$\evalfunset_{\accfun}=\lossset\cup\{\accfun\}$.
Thus, $\evalfunset_{\accfun}$ contains all the considered loss-based criteria and the target metric $\accfun$ itself, while excluding the other metrics in $\accset$.
In other words, checkpoint selection was based either on one of the validation losses or directly on the target validation metric. Including metrics other than $\accfun$ would introduce an additional metric--metric mismatch and would address a different question, namely whether one evaluation metric can serve as a proxy for another. In all cases, checkpoint selection relied exclusively on the validation set, whereas the test set was used only for retrospective evaluation. 

Summarizing, for each training loss $\lossfun_{\mathrm{train}}\in\lossset$, for each validation criterion $v\in\evalfunset_\accfun$ and selection protocol $q\in\mathcal{Q}=\{\mathrm{PH},\mathrm{ES}\}$, we computed:
\begin{enumerate}
    \item the test performance of the checkpoint selected under protocol $q$ using $v$, i.e. $m_{Test}^{q}(v)=m\!\left(Test,\theta_{\hat e_{v,Val}^{\,q}}\right);$
    \item the best-observed value of the same target metric along the complete training trajectory, i.e. $m_{Test}^{\star}=\max_{e\in\epochset} m(Test,\theta_e).$
\end{enumerate}

\end{insertedtext}

\begin{figure}
    \centering
    \begin{tikzpicture}

\definecolor{darkgray176}{RGB}{176,176,176}
\definecolor{darkorange25512714}{RGB}{255,127,14}
\definecolor{forestgreen4416044}{RGB}{44,160,44}
\definecolor{lightgray204}{RGB}{204,204,204}
\definecolor{steelblue31119180}{RGB}{31,119,180}


\begin{axis}[
scale only axis,
width=0.90\linewidth,
height=0.52\linewidth,
axis y line*=left,
axis x line*=bottom,
legend cell align={left},
legend style={
    at={(0.98,0.02)},
    anchor=south east,
    fill opacity=0.8,
    draw opacity=1,
    text opacity=1,
    draw=lightgray204,
    font=\scriptsize
},
tick align=outside,
tick pos=left,
x grid style={darkgray176},
y grid style={darkgray176},
xmajorgrids,
ymajorgrids,
xlabel={Epoch $t$},
xlabel style={yshift=-3.5ex},
xmin=1,
xmax=60,
xtick={27,33,44,60},
xticklabels={
    $\hat{t}_{\mathrm{PH}}(\accfun)$,
    $\hat{t}_{\mathrm{PH}}(\lossfun)$,
    $t^\star_{\accfun,Test}$,
    $E$
},
xticklabel style={
    font=\scriptsize,
    rotate=25,
    anchor=north east
},
ylabel={Raw validation loss $\lossfun(Val,\theta_t)$},
ymin=0.228035795738574,
ymax=1.92603409331468,
ytick style={color=black}
]

\addplot [
    semithick,
    steelblue31119180
]
table {%
1 1.84885235251576
2 1.77957445515552
3 1.70278859213896
4 1.62415747079044
5 1.55194571237138
6 1.46995793310122
7 1.39920320069884
8 1.32976019041497
9 1.23619595669576
10 1.15851172310106
11 1.09444895160927
12 1.02544040565383
13 0.961438877614752
14 0.897625221270278
15 0.855773176007305
16 0.818470039894834
17 0.764072898022399
18 0.736835527342369
19 0.69327176097321
20 0.667038806542299
21 0.631140924047725
22 0.611255221883297
23 0.568544211651967
24 0.544741185539857
25 0.506608094607217
26 0.46653194463008
27 0.411771242454243
28 0.394154943103816
29 0.368778415977071
30 0.346188282959007
31 0.315496512228786
32 0.313183520156312
33 0.305217536537488
34 0.308840865524703
35 0.331303756313516
36 0.327793123513309
37 0.347891934489244
38 0.370132681262531
39 0.373036080367451
40 0.39027791599238
41 0.403871355730213
42 0.413701512488994
43 0.412521316231385
44 0.431961473876863
45 0.452440334839384
46 0.442038611061614
47 0.478185062740618
48 0.494429929419585
49 0.51665415264276
50 0.572664166722595
51 0.604106514346459
52 0.635641763802065
53 0.697915428400534
54 0.757537275059337
55 0.809054579113322
56 0.874248948542963
57 0.925799852793272
58 0.987640647190179
59 1.04776577006701
60 1.0828646775505
};

\addlegendentry{$\lossfun(Val,\theta_t)$}

\addplot [
    line width=0.55pt,
    darkorange25512714,
    dash pattern=on 4.44pt off 1.92pt,
    forget plot
]
coordinates {
    (27,0.228035795738574)
    (27,1.92603409331468)
};

\addplot [
    line width=0.55pt,
    steelblue31119180,
    dash pattern=on 4.44pt off 1.92pt,
    forget plot
]
coordinates {
    (33,0.228035795738574)
    (33,1.92603409331468)
};

\addplot [
    line width=0.55pt,
    forestgreen4416044,
    dash pattern=on 4.44pt off 1.92pt,
    forget plot
]
coordinates {
    (44,0.228035795738574)
    (44,1.92603409331468)
};

\addplot [
    only marks,
    mark=*,
    mark size=2.2pt,
    steelblue31119180,
    forget plot
]
coordinates {
    (33,0.305217536537488)
};

\draw[->,draw=black]
(axis cs:25,0.43) --
(axis cs:33,0.305217536537488);

\draw (axis cs:25,0.43) node[
    scale=0.55,
    anchor=south east,
    text=black
]{
    $\lossfun(Val,\theta_{\hat{t}_{\mathrm{PH}}(\lossfun)})$
};

\end{axis}


\begin{axis}[
scale only axis,
width=0.90\linewidth,
height=0.52\linewidth,
axis y line*=right,
axis x line=none,
xmin=1,
xmax=60,
xtick=\empty,
ylabel={
    Validation and test metric
},
ymin=0.48,
ymax=1.02,
ytick pos=right,
ytick style={color=black},
yticklabel style={anchor=west},
legend cell align={left},
legend style={
    at={(0.98,0.12)},
    anchor=south east,
    fill opacity=0.8,
    draw opacity=1,
    text opacity=1,
    draw=lightgray204,
    font=\scriptsize
}
]

\addplot [
    semithick,
    darkorange25512714
]
table {%
1 0.55000203354798
2 0.56795688947098
3 0.587861124345703
4 0.608251717506668
5 0.626995559884497
6 0.648305102787675
7 0.666764073181447
8 0.684989731019167
9 0.709645542121454
10 0.7304743202558
11 0.748211328299691
12 0.767860404033209
13 0.787067981714917
14 0.807317021350912
15 0.823167158445876
16 0.839244557713061
17 0.861122319312232
18 0.87709559486966
19 0.897966486969893
20 0.914333667124724
21 0.932365530761859
22 0.944546176639309
23 0.960180409049454
24 0.967932561115391
25 0.976229831206458
26 0.982048123538947
27 0.989210499357443
28 0.985045693211118
29 0.982052008140334
30 0.978323486425225
31 0.977361310853515
32 0.970178769485933
33 0.965831895926689
34 0.959887914323951
35 0.950354042583266
36 0.94864079973786
37 0.94166563741267
38 0.934765914012937
39 0.933314992277183
40 0.92843650086326
41 0.924683066628662
42 0.922011399038097
43 0.922253061035676
44 0.917183852120968
45 0.911862175463837
46 0.914550741915381
47 0.905181192656438
48 0.900970443159808
49 0.895211014056373
50 0.880697024310285
51 0.872549341079779
52 0.864377629734945
53 0.848240730904437
54 0.832790999752621
55 0.819441389641615
56 0.802547660697655
57 0.78918934459773
58 0.773164623248882
59 0.757584481553331
60 0.748489349174735
};

\addlegendentry{$\accfun(Val,\theta_t)$}

\addplot [
    semithick,
    forestgreen4416044
]
table {%
1 0.501015693052257
2 0.516532944255646
3 0.54037811175728
4 0.55520015188338
5 0.577887358497374
6 0.602110537979375
7 0.626836644146879
8 0.646963341801202
9 0.660077024887424
10 0.683858173393926
11 0.708496428079538
12 0.714076011623153
13 0.739136131806784
14 0.7583283139484
15 0.776487196367053
16 0.783798895462689
17 0.793516743273173
18 0.800721425510152
19 0.81316766281778
20 0.829176497269885
21 0.829191176229134
22 0.835233438190647
23 0.85015434906982
24 0.854297568545418
25 0.864349462536276
26 0.870770376112029
27 0.89134941990354
28 0.894250092434907
29 0.909620452817627
30 0.913831843806428
31 0.919718742045028
32 0.925118191555489
33 0.924038718931387
34 0.932394895034604
35 0.924051558864345
36 0.931735563944225
37 0.921413614900481
38 0.92873769542988
39 0.923486702593883
40 0.923071534841301
41 0.934034280148708
42 0.934027735029954
43 0.944574320817911
44 0.953496217133108
45 0.934956684472344
46 0.939653309272614
47 0.933137563799111
48 0.927740765830905
49 0.914539035545872
50 0.89424635516931
51 0.884587980472821
52 0.869087875971693
53 0.838364521124796
54 0.821201379298286
55 0.80256904103081
56 0.774863753041795
57 0.763624235123653
58 0.738153189011995
59 0.728562105858612
60 0.713321257913023
};

\addlegendentry{$\accfun(Test,\theta_t)$}

\addplot [
    only marks,
    mark=*,
    mark size=2.2pt,
    darkorange25512714,
    forget plot
]
coordinates {
    (27,0.989210499357443)
};

\addplot [
    only marks,
    mark=*,
    mark size=2.2pt,
    darkorange25512714,
    forget plot
]
coordinates {
    (27,0.89134941990354)
};

\addplot [
    only marks,
    mark=*,
    mark size=2.2pt,
    steelblue31119180,
    forget plot
]
coordinates {
    (33,0.924038718931387)
};

\addplot [
    only marks,
    mark=star,
    mark size=3pt,
    forestgreen4416044,
    forget plot
]
coordinates {
    (44,0.953496217133108)
};

\addplot [
    line width=0.4pt,
    darkorange25512714,
    opacity=0.6,
    dotted,
    forget plot
]
coordinates {
    (27,0.89134941990354)
    (60,0.89134941990354)
};

\addplot [
    line width=0.4pt,
    steelblue31119180,
    opacity=0.6,
    dotted,
    forget plot
]
coordinates {
    (33,0.924038718931387)
    (60,0.924038718931387)
};

\addplot [
    line width=0.4pt,
    forestgreen4416044,
    opacity=0.6,
    dotted,
    forget plot
]
coordinates {
    (44,0.953496217133108)
    (60,0.953496217133108)
};

\draw[->,draw=black]
(axis cs:21,0.855) --
(axis cs:27,0.89134941990354);

\draw (axis cs:21,0.855) node[
    scale=0.52,
    anchor=north east,
    text=black
]{
    $\accfun(Test,\theta_{\hat{t}_{\mathrm{PH}}(\accfun)})$
};

\draw[->,draw=black]
(axis cs:37,0.895) --
(axis cs:33,0.924038718931387);

\draw (axis cs:37,0.895) node[
    scale=0.52,
    anchor=north west,
    text=black
]{
    $\accfun(Test,\theta_{\hat{t}_{\mathrm{PH}}(\lossfun)})$
};

\draw[->,draw=black]
(axis cs:48,0.980) --
(axis cs:44,0.953496217133108);

\draw (axis cs:48,0.980) node[
    scale=0.52,
    anchor=south west,
    text=black
]{
    $\accfun^\star_{Test}$
};

\end{axis}

\end{tikzpicture}
    \caption{Illustration of post-hoc checkpoint selection using validation loss and the target validation metric. The raw validation loss is shown on the left axis, whereas validation and test metric values are shown on the right axis. The figure displays the validation loss in its original minimization form. Metric-based validation selection retains checkpoint $\theta_{\hat{t}_{\mathrm{PH}}(\accfun)}$, whereas loss-based selection retains checkpoint $\theta_{\hat{t}_{\mathrm{PH}}(\lossfun)}$. Their test performances are compared retrospectively with $\accfun^\star_{Test}=\max\limits_{t\in\epochset}\accfun(Test,\theta_t)$. Test-set information is shown only to quantify the resulting checkpoint-selection gaps and is not used to select either checkpoint.}
    \label{fig:ex3}
\end{figure}

\begin{insertedtext}
    
For each experimental run, target metric ${\accfun}\in\accset$, validation criterion $v\in\evalfunset_{\accfun}$, and selection protocol $q\in\mathcal{Q}$, we define the \textit{relative checkpoint-selection gap} as $$ \Delta_{\accfun}^q(v)= \frac{ {\accfun}_{Test}^{\star}-m_{Test}^{q}(v)}{{\accfun}_{Test}^{\star}}.$$  

The resulting gaps were collected across datasets, data splits, and, where applicable, parameter-to-sample ratios.
\end{insertedtext}

\section{Experimental assessment}
\label{sec:exp}
\subsection{Datasets}

\begin{insertedtext}
The experimental evaluation covers two data types: 26 tabular classification datasets from the UCI Machine Learning Repository~\citep{asuncion2007uci}, and two image classification benchmarks, CIFAR-10~\citep{krizhevsky2009learning} and TinyImageNet-200~\citep{le2015tiny}. The tabular experiments allow us to examine checkpoint selection across datasets with different sample sizes, numbers of classes, and degrees of class separability. The image benchmarks are included to assess whether the observed selection behaviour extends to high-dimensional inputs and deep neural-network architectures.
\end{insertedtext}

\paragraph{Tabular datasets.}
\label{subsec:tabular}
The UCI datasets considered in the study are listed in Table~\ref{tab:datasets}, together with their numbers of instances and target classes. 
\begin{insertedtext}
Datasets were selected among those containing at least 1,000 instances.
\end{insertedtext}

They were processed using the same pipeline to avoid introducing dataset-specific modelling choices. 
Categorical and binary variables were represented through one-hot encoding, whereas the remaining numerical variables were retained as numerical features. 
After encoding, input features were standardized using z-score normalization. For each cross-validation fold, the normalization parameters were estimated exclusively from the training partition and then applied unchanged to the corresponding validation and test partitions. No dataset-specific feature engineering or hyperparameter optimization was performed.
For the tabular experiments, we additionally characterize class separability using the generalized discrimination value (GDV)~\citep{SCHILLING2021278}. The GDV is used as a descriptive measure to order the UCI datasets and to examine whether checkpoint-selection behaviour varies with class separability. It is not used for training, checkpoint selection, or hyperparameter tuning. Since the GDV depends on the adopted input representation, we do not use it to make direct separability comparisons between the tabular and image benchmarks. 

\paragraph{Image datasets.}
\begin{insertedtext}
CIFAR-10 consists of 60,000 RGB images of size $32\times32$ belonging to 10 classes, whereas TinyImageNet-200 contains 120,000 RGB images of size $64\times64$ belonging to 200 classes. These datasets were evaluated using the deep architecture described in Section~\ref{sec:models}. For both benchmarks, the training, validation, and test partitions used in the experiments were mutually disjoint, and the validation partition was used exclusively for checkpoint selection.
\end{insertedtext}

\begin{table}[t]
\centering

\scalebox{0.7}{%
\begin{tabular}{lcc|lcc}
\toprule
\textbf{Name} & \textbf{Instances} & \textbf{N. classes} &
\textbf{Name} & \textbf{Instances} & \textbf{N. classes} \\
\midrule
Pen-Based Recognition of Handwritten Digits & 10992 & 10 & Cardiotocography & 2126 & 10 \\
Page Blocks Classification & 5473 & 5 & Maternal Health Risk & 1013 & 3\\
Molecular Biology (Splice-junction Gene Sequences) & 3190 & 3 & Spambase & 4601 & 2\\
Steel Plates Faults & 1941 & 7 & Bank Marketing & 45211 & 2\\
Website Phishing & 1353 & 3 & Letter Recognition & 20000 & 26\\
Taiwanese Bankruptcy Prediction & 6819 & 2 & Waveform Database Generator (Version 1) & 5000 & 3\\
Statlog (Image Segmentation) & 2310 & 7 & Optical Recognition of Handwritten Digits & 5620 & 10 \\
Drug Consumption (Quantified) & 1885 & 7 & Yeast & 1484 & 10 \\
Contraceptive Method Choice & 1473 & 3 & Hepatitis C Virus (HCV) for Egyptian patients & 1385 & 4\\
Room Occupancy Estimation & 10129 & 4 & Statlog (German Credit Data) & 1000 & 2\\
ISOLET & 7797 & 26 & Chess (King-Rook vs. King-Pawn) & 3196 & 2\\
Musk (Version 2) & 6598 & 2 & Predict Students' Dropout and Academic Success & 4424 & 3\\
NHANES 2013-2014 Age Prediction Subset & 6287 & 2 & Polish Companies Bankruptcy & 10503 & 2 \\
\bottomrule
\end{tabular}
}
\caption{
Summary of the datasets used in the experimental evaluation 
retrieved from the UCI Machine Learning Repository~\citep{asuncion2007uci}. For each dataset, we report the total number of instances and the number of target classes.}
\label{tab:datasets}
\end{table}

\subsection{Models}
\label{sec:models}

We consider two complementary architectural settings. The tabular experiments use a shallow fully connected network whose size can be systematically controlled through the parameter-to-sample ratio introduced below. The image experiments instead use a standard deep convolutional architecture to examine whether the observed checkpoint-selection behaviour persists beyond the controlled shallow setting.

\paragraph{Tabular setting.}
For the UCI datasets, we employ fully connected neural networks with a single hidden layer, ReLU activations, and an output layer producing $K$ class logits. This architecture provides a controlled setting in which model capacity can be varied while limiting the additional optimization effects introduced by depth.

Following the insights of \citep{advani2020high}, we employed fully connected neural networks with a single hidden layer and ReLU activation functions, in order to preserve architectural simplicity and experimental controllability while exploring different generalization regimes. Indeed, as shown in \citep{advani2020high}, generalization behavior is related to the ratio between the number of trainable parameters and the number of training samples. Accordingly, let $n_\theta$ denote the total number of trainable model parameters and let $N_{\mathrm{Train}}$ denote the number of observations used for parameter optimization. We define the parameter-to-sample ratio as
$r=n_\theta/N_{\mathrm{Train}}$. This ratio is the reciprocal of the sample-to-parameter ratio $N_{\mathrm{Train}}/n_\theta$ adopted by \citet{advani2020high}.
Accordingly, $r = 1$ corresponds to an equal number of model parameters and samples, while values below or above $1$ indicate under- and over-parameterized regimes, respectively. Thus, the use of a shallow architecture allows us to systematically explore these regimes by varying the number of hidden units so as to control the total number of trainable parameters relative to the size of the training dataset. In our experiments, we consider the values $r \in \{0.3, 0.5, 0.7, 0.8, 1, 1.2, 5, 10, 50\}$.
Notice that we deliberately focus on shallow neural networks with a single hidden layer, as our goal is not to achieve state-of-the-art performance, but to isolate and analyze the effect of validation criteria on model selection. Deeper architectures introduce multiple additional factors--such as hierarchical representations, layer-wise implicit regularization, and complex optimization dynamics--that can confound the interpretation of validation-based checkpoint selection protocols. By adopting a controlled shallow setting, we are able to systematically vary the parameter-to-sample ratio and explore different generalization regimes while keeping architectural and optimization-related effects to a minimum. This choice enables a clearer assessment of how different validation criteria influence model selection, independently of depth-related phenomena.

\paragraph{Image setting.}
\begin{insertedtext}
For CIFAR-10 and TinyImageNet-200, we employ a ResNet-50 architecture~\citep{he2016deep}, replacing its final classification layer to match the number of classes in each dataset. In contrast to the controlled ratio between  number of model parameters and samples used for the tabular datasets, the image experiments use a fixed deep architecture. Their purpose is to assess whether the conclusions concerning validation-based checkpoint selection extend to a substantially different data modality and architectural regime. All experiments were conducted using fixed random seeds to control stochastic factors between different runs, including data partitioning, model initialization, and data shuffling. The image models were trained from scratch, without pretrained weights or data augmentation. The only preprocessing operation was channel-wise z-score normalization. For each data partition, the channel means and standard deviations were estimated exclusively from the portion used for parameter optimization and were subsequently applied to the corresponding validation and test sets.
\end{insertedtext}

\subsection{Adopted losses}
We consider $\lossset=\{\textbf{CE}, \textbf{PolyLoss}, \textbf{C-Loss}\}$ corresponding respectively to cross-entropy, PolyLoss and C-Loss as loss functions.
The losses described below are reported in their conventional minimization form and are used in this form for model training.

The cross-entropy loss, widely used in classification tasks, emerges naturally from the principle of maximum likelihood estimation under the assumption that the model outputs a categorical distribution over the classes. It is defined as 
$$\lossfun_{\text{CE}} = - \sum_{i=1}^N \sum_{k=1}^K t_k^{(i)} \log \left( m_k^{(i)} \right)$$
where $\vec{t}$ is the one-hot encoded target vector \( \vec{t}^{(i)} \in \{0,1\}^K \) of the actual label $y^{(i)}$, i.e. \( t_k^{(i)} = 1 \) if \( k = y^{(i)} \), and \( t_k^{(i)} = 0 \) otherwise, and \( \vec{m}^{(i)} = (m_1^{(i)}, \dots, m_K^{(i)}) \) is the class output probability distribution of the model $M$ on the input $\vec{x}^{(i)}$.

In \citep{leng2022polyloss} was introduced the \emph{PolyLoss}, a polynomial reparameterization of cross-entropy obtained via its Taylor expansion around the correct-class confidence. Denoting by $p_{y^{(i)}}$ the predicted probability for the true class of sample $i$, the loss takes the form
$\lossfun_{\text{PolyLoss}} \;=\; \sum_{j=1}^{\infty} \alpha_j \,\big(1 - p_{y^{(i)}}\big)^j,$
with coefficients $\{\alpha_j\}_{j\ge 1}$ to be tuned. In its natural (infinite) form, PolyLoss is impractical and  does not consistently outperform standard cross-entropy. To address this, the authors propose a simplified, first-order truncation,
$\lossfun_{\text{Poly-1}} \;=\; -\log p_{y^{(i)}} \;+\; \epsilon\,\big(1 - p_{y^{(i)}}\big),$ controlled by a scalar hyperparameter $\epsilon$.
In this work, Poly-1 was configured with $\epsilon = 1$.

Finally, the C-Loss \citep{singh2010loss,singh2014c} is a surrogate for the 0–1 loss built from the \emph{correntropy} \citep{santamaria2006generalized} between true labels and model scores. Unlike cross-entropy, the C-Loss can be more robust to outliers and label noise. In binary classification problems where $y^{(i)}\in \{-1,1\}$ and single output $p^{(i)}=M(\vec{x}^{(i)})$, it is defined via a positive-definite kernel $\mathcal{k}(\cdot)$ (typically Gaussian):
$$\lossfun_{{C}}(y^{(i)},p^{(i)})\;= \beta\big(1-\mathcal{k}_\sigma(y^{(i)}-p^{(i)})\big)$$
with $\mathcal{k}_\sigma(u)=\exp\!\Big(-\tfrac{u^{2}}{2\sigma^{2}}\Big)$, $\beta$ and $\sigma$ parameters properly chosen. Multiclass variants can be built by applying the one-class-versus-the-rest strategy.
In this work, C-Loss was used with parameters $\sigma = 0.5$ and $\beta = 1$. 

\begin{insertedtext}
\subsection{Adopted evaluation metrics}
\begin{insertedtext}
We consider $\accset=\{\textbf{acc}, \textbf{mF1}, \textbf{MCC}\}$ corresponding respectively to accuracy, macro-averaged F1 score (macro-F1), and Matthews correlation coefficient (MCC) as target evaluation metrics.    
\end{insertedtext}
Given an input $\mathbf{x}^{(i)}$, the predicted class is $\hat{y}^{(i)}=\arg\max_{1\leq k\leq K} p_k^{(i)},$ where $p_k^{(i)}$ denotes the probability assigned by the model to class $k$. 
Accuracy measures the proportion of correctly classified instances and is defined as ${\mathrm{Acc}}=\frac{1}{N}\sum_{i=1}^{N}\mathbf{1}\left[\hat{y}^{(i)}=y^{(i)}\right],$ where $\mathbf{1}[\cdot]$ denotes the indicator function.

Macro-F1 assigns the same importance to every class, independently of its frequency. For each class $k$, considered in a one-versus-rest fashion, let $\mathrm{TP}_k$, $\mathrm{FP}_k$, and $\mathrm{FN}_k$ denote the numbers of true positives, false positives, and false negatives, respectively. The class-specific F1 score is $\mathrm{F1}_k=\frac{2\mathrm{TP}_k}{2\mathrm{TP}_k+\mathrm{FP}_k+\mathrm{FN}_k},$
and macro-F1 is obtained by averaging across the $K$ classes: ${\mathrm{mF1}}=\frac{1}{K}\sum_{k=1}^{K}\mathrm{F1}_k.$ When the denominator of a class-specific F1 score is zero, that score is set to zero.

MCC provides a correlation-based measure that takes into account all entries of the confusion matrix and is particularly informative in the presence of class imbalance. Let $\mathbf{C}\in\mathbb{N}^{K\times K}$ be the confusion matrix, where $C_{ij}$ is the number of instances whose true class is $i$ and whose predicted class is $j$. We define $s=\sum_{i=1}^{K}\sum_{j=1}^{K}C_{ij},\ c=\sum_{k=1}^{K}C_{kk},$
$t_k=\sum_{j=1}^{K}C_{kj},\ p_k=\sum_{i=1}^{K}C_{ik},$
where $t_k$ and $p_k$ are, respectively, the numbers of instances belonging to and predicted as class $k$. The multiclass MCC is defined as $\mathrm{MCC}=\frac{cs-\sum_{k=1}^{K}p_k t_k}{\sqrt{\left(s^2-\sum_{k=1}^{K}p_k^2\right)\left(s^2-\sum_{k=1}^{K}t_k^2\right)}}.$

When the denominator is zero, MCC is set to zero. Since the standard MCC takes values in $[-1,1]$, we rescale it to $[0,1]$ so that it has the same numerical range as the other target metrics
considered in this study.

Accordingly, all the adopted target metrics take values in $[0,1]$, with larger values indicating better predictive performance. 

For each target metric $m\in\accset$, the same metric is computed on both the validation and test sets. 
\end{insertedtext}

\subsection{Training and Evaluation Protocol}
Models for tabular datasets were trained for a maximum of $E=20,000$ epochs for each dataset using stochastic gradient descent with a batch size of $64$ samples. The learning rate was set to $0.01$ and fixed through all the training epochs. 
All results were computed under a $10$-fold stratified cross-validation~\citep{bishop2023deep} scheme.
For each fold, the \SI{15}{\percent} of the training set was used for validation set $Val$ using stratified sampling
~\citep{bishop2023deep}. 

Prior to each training, all input features were then standardized using $z$-score normalization~\citep{hastie2009elements,apicella2023effects}. The normalization parameters (mean and standard deviation) were computed exclusively on the training portion of each fold and subsequently applied to the corresponding validation and test sets, ensuring that no information from the held-out data leaked into the training process \citep{apicella2025don}. 

We emphasize that, as above discussed, the objective of this work is not to achieve state-of-the-art performance on these benchmarks, thus we intentionally adopt simple and uniform preprocessing rather than dataset-specific preprocessing prior to each training.

Models were trained in separate runs, each using a single loss function $\lossfun \in \lossset$ as the training objective. For both protocols, each model was trained for the complete predefined number of epochs and the per-epoch validation and test values were recorded. Patience-based early stopping was then reconstructed retrospectively from the validation sequence: only checkpoints up to the resulting stopping epoch were eligible for selection.
Checkpoints following that epoch were used exclusively to quantify whether termination prevented the procedure from reaching a better-performing checkpoint and were never available to the early-stopping rule.
\begin{insertedtext}
In case of patience-based early stopping, we set the patience $T=20$.
The empirical distributions of the relative checkpoint-selection gaps are visualized using strip plots. For the tabular experiments, each point represents an individual experimental run, datasets are shown on the vertical axis, and colors identify the parameter-to-sample ratio $r$. Separate figures are reported for each training loss, target metric, and selection protocol. For the image experiments, points are
grouped by dataset, while marker shapes distinguish the training objectives.

To complement these detailed visualizations,  we report the 95th percentile of the gap distribution for each combination of data modality, training loss, validation criterion, target metric, and selection protocol -- the lower, the better, as low values indicate small gaps while long tails point to severe ones. Thus, the strip plots expose the heterogeneity across individual datasets and configurations, whereas the 95th percentile summarizes the magnitude of the more severe checkpoint-selection gaps.
\end{insertedtext}

\section{Results} 
\label{sec:results}

\begin{table}[htbp]
\centering
\caption{Empirical 95th percentile of the relative checkpoint-selection gap under post-hoc checkpoint selection, for each combination of target metric, dataset type, training loss, and validation criterion. Lower values indicate smaller upper-tail gaps and are therefore preferable. In boldface the best values.}
\label{tab:selection_gap}
\begin{tabular}{l|l|l||rrrr}
\toprule
\textbf{Metric} & \textbf{Dataset Type} & \textbf{Training Loss} & \multicolumn{4}{c}{\textbf{Val. Criterion}} \\
 & & & CE & CLoss & PolyLoss & Metric \\
\midrule
\multirow{6}{*}{MCC}
 & \multirow{3}{*}{Image} & CE      & 0.083 & 0.005 & 0.069 & \textbf{0.005} \\
 &                        & CLoss   & 0.055 & 0.011 & 0.050 & \textbf{0.007} \\
 &                        & PolyLoss& 0.096 & \textbf{0.005} & 0.095 & 0.006 \\
\cmidrule(l){2-7}
 & \multirow{3}{*}{UCI}   & CE      & 0.165 & 0.129 & 0.159 & \textbf{0.107} \\
 &                        & CLoss   & 0.272 & 0.095 & 0.217 & \textbf{0.090} \\
 &                        & PolyLoss& 0.177 & 0.145 & 0.178 & \textbf{0.107} \\
\midrule
\multirow{6}{*}{Accuracy}
 & \multirow{3}{*}{Image} & CE      & 0.183 & \textbf{0.016} & 0.155 & 0.018 \\
 &                        & CLoss   & 0.432 & 0.075 & 0.400 & \textbf{0.051} \\
 &                        & PolyLoss& 0.206 & \textbf{0.013} & 0.201 & 0.017 \\
\cmidrule(l){2-7}
 & \multirow{3}{*}{UCI}   & CE      & 0.125 & 0.193 & 0.105 & \textbf{0.104} \\
 &                        & CLoss   & 0.363 & \textbf{0.072} & 0.277 & 0.078 \\
 &                        & PolyLoss& 0.127 & 0.195 & 0.114 & \textbf{0.107} \\
\midrule
\multirow{6}{*}{macro-F1}
 & \multirow{3}{*}{Image} & CE      & 0.210 & \textbf{0.014} & 0.175 & 0.021 \\
 &                        & CLoss   & 0.661 & 0.246 & 0.639 & \textbf{0.037} \\
 &                        & PolyLoss& 0.216 & \textbf{0.015} & 0.201 & 0.015 \\
\cmidrule(l){2-7}
 & \multirow{3}{*}{UCI}   & CE      & 0.312 & 0.218 & 0.312 & \textbf{0.171} \\
 &                        & CLoss   & 0.566 & 0.221 & 0.502 & \textbf{0.125} \\
 &                        & PolyLoss& 0.324 & 0.257 & 0.328 & \textbf{0.178} \\
\bottomrule
\end{tabular}
\end{table}

\begin{table}[htbp]
\centering
\caption{Empirical 95th percentile of the relative checkpoint-selection gap under patience-based early stopping, for each combination of target metric, dataset type, training loss, and validation criterion. Lower values indicate smaller upper-tail gaps and are therefore preferable. In boldface the best values.} 
\label{tab:selection_gap_es_p95}
\begin{tabular}{l|l|l||rrrr}
\toprule
\textbf{Metric} & \textbf{Dataset Type} & \textbf{Training Loss} & \multicolumn{4}{c}{\textbf{Val. Criterion}} \\
 & & & CE & CLoss & PolyLoss & Metric \\
\midrule
\multirow{6}{*}{MCC}
 & \multirow{3}{*}{Image} & CE      & 0.083 & \textbf{0.005} & 0.069 & 0.007 \\
 &                        & CLoss   & 0.346 & 0.025 & 0.346 & \textbf{0.018} \\
 &                        & PolyLoss& 0.096 & \textbf{0.005} & 0.095 & 0.007 \\
\cmidrule(l){2-7}
 & \multirow{3}{*}{UCI}   & CE      & \textbf{0.185} & 0.279 & 0.187 & 0.324 \\
 &                        & CLoss   & 0.395 & \textbf{0.201} & 0.393 & 0.382 \\
 &                        & PolyLoss& 0.211 & 0.321 & \textbf{0.197} & 0.334 \\
\midrule
\multirow{6}{*}{Accuracy}
 & \multirow{3}{*}{Image} & CE      & 0.183 & \textbf{0.016} & 0.155 & 0.030 \\
 &                        & CLoss   & 0.904 & 0.205 & 0.904 & \textbf{0.140} \\
 &                        & PolyLoss& 0.206 & \textbf{0.013} & 0.201 & 0.018 \\
\cmidrule(l){2-7}
 & \multirow{3}{*}{UCI}   & CE      & 0.171 & \textbf{0.163} & 0.171 & 0.198 \\
 &                        & CLoss   & 0.475 & \textbf{0.180} & 0.447 & 0.477 \\
 &                        & PolyLoss& 0.171 & \textbf{0.169} & 0.204 & 0.170 \\
\midrule
\multirow{6}{*}{macro-F1}
 & \multirow{3}{*}{Image} & CE      & 0.210 & \textbf{0.014} & 0.175 & 0.021 \\
 &                        & CLoss   & 0.965 & \textbf{0.443} & 0.965 & 0.972 \\
 &                        & PolyLoss & 0.216 & \textbf{0.015} & 0.201 & 0.017 \\
\cmidrule(l){2-7}
 & \multirow{3}{*}{UCI}   & CE      & \textbf{0.335} & 0.387 & 0.337 & 0.434 \\
 &                        & CLoss   & 0.683 & \textbf{0.444} & 0.643 & 0.690 \\
 &                        & PolyLoss& 0.350 & 0.424 & \textbf{0.345} & 0.440 \\
\bottomrule
\end{tabular}
\end{table}

\begin{sidewaysfigure}[p]
\centering

\begin{subfigure}[t]{0.96\textheight}
    \centering
    \includegraphics[
        width=\linewidth,
        height=0.26\textwidth,
        keepaspectratio
    ]{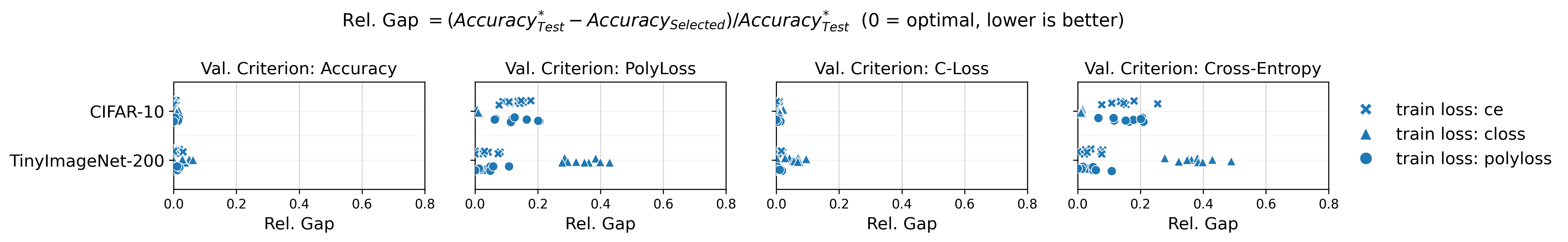}
    \caption{Accuracy.}
    \label{fig:image_acc_ph}
\end{subfigure}

\vspace{0.3em}

\begin{subfigure}[t]{0.96\textheight}
    \centering
    \includegraphics[
        width=\linewidth,
        height=0.26\textwidth,
        keepaspectratio
    ]{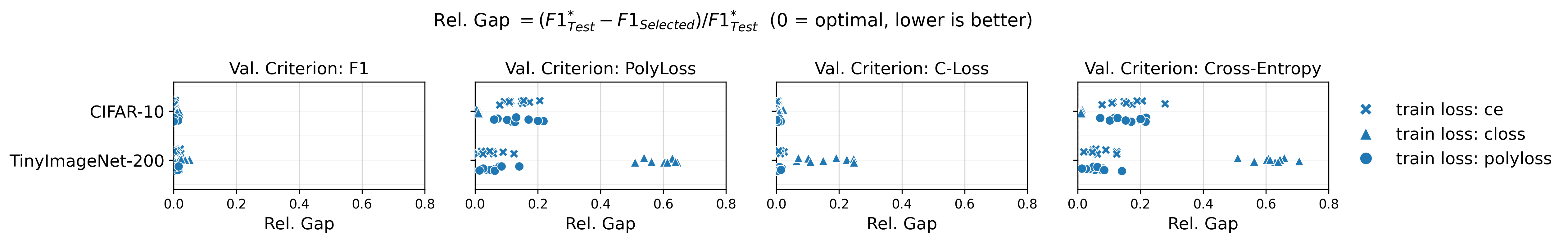}
    \caption{Macro-F1.}
    \label{fig:image_f1_ph}
\end{subfigure}

\vspace{0.3em}

\begin{subfigure}[t]{0.96\textheight}
    \centering
    \includegraphics[
        width=\linewidth,
        height=0.26\textwidth,
        keepaspectratio
    ]{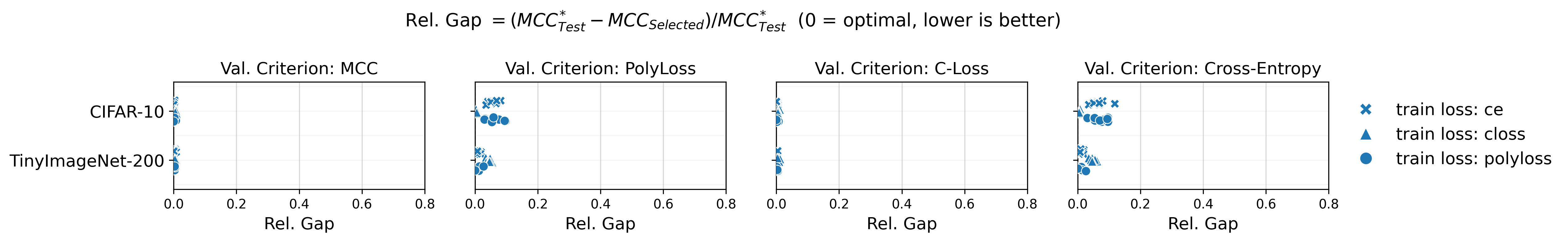}
    \caption{MCC.}
    \label{fig:image_mcc_ph}
\end{subfigure}

\caption{Relative checkpoint-selection gaps obtained on the image datasets
under post-hoc checkpoint selection. From top to bottom, the target metrics
are accuracy, macro-F1, and MCC. Within each panel, columns correspond to
the considered validation criteria, while marker shapes identify the loss
used to train the model. A value of zero indicates that the selected
checkpoint attains the best-observed test performance over the complete
predefined training run; lower values are preferable.}

\label{fig:image_posthoc}
\end{sidewaysfigure}

\begin{sidewaysfigure}[p]
\centering

\begin{subfigure}[t]{0.96\textheight}
    \centering
    \includegraphics[
        width=\linewidth,
        height=0.26\textwidth,
        keepaspectratio
    ]{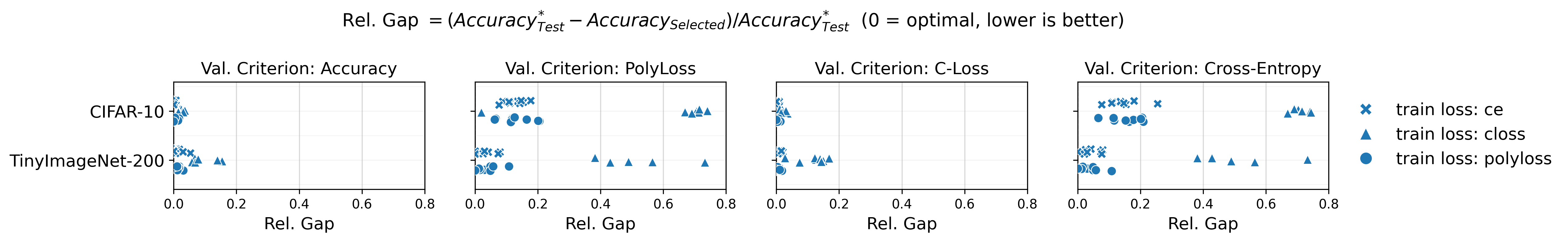}
    \caption{Accuracy.}
    \label{fig:image_acc_es}
\end{subfigure}

\vspace{0.3em}

\begin{subfigure}[t]{0.96\textheight}
    \centering
    \includegraphics[
        width=\linewidth,
        height=0.26\textwidth,
        keepaspectratio
    ]{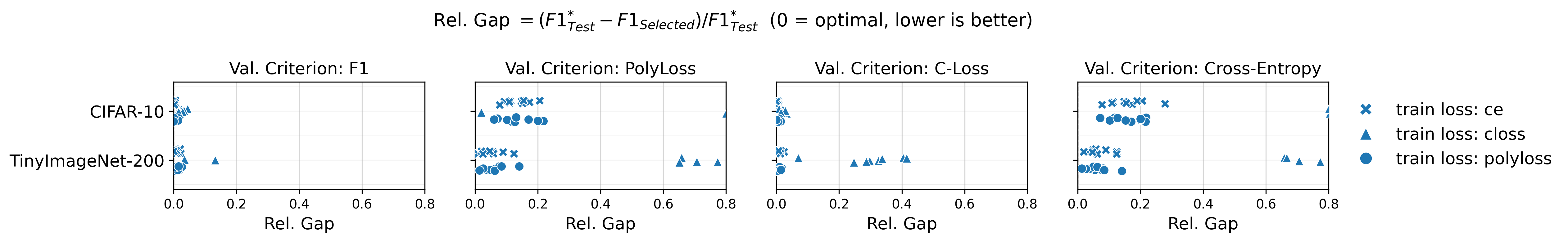}
    \caption{Macro-F1.}
    \label{fig:image_f1_es}
\end{subfigure}

\vspace{0.3em}

\begin{subfigure}[t]{0.96\textheight}
    \centering
    \includegraphics[
        width=\linewidth,
        height=0.26\textwidth,
        keepaspectratio
    ]{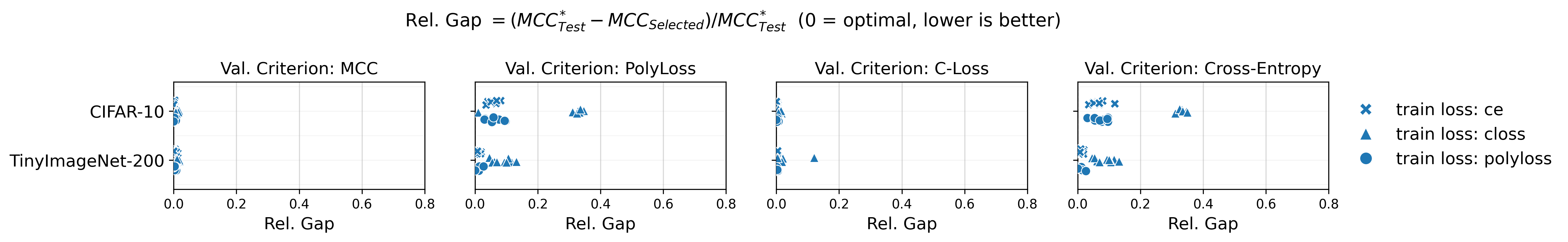}
    \caption{MCC.}
    \label{fig:image_mcc_es}
\end{subfigure}

\caption{Relative checkpoint-selection gaps obtained on the image datasets under patience-based early stopping with $T=20$. From top to bottom, the target metrics are accuracy, macro-F1, and MCC. Within each panel, columns
correspond to the considered validation criteria, while marker shapes identify the loss used to train the model. A value of zero indicates no gap with respect to the best-observed test performance over the complete predefined training run; lower values are preferable. The reported gap may reflect both checkpoint ranking and termination before better-performing checkpoints are reached.}

\label{fig:image_earlystopping}
\end{sidewaysfigure}

\begin{insertedtext}

\subsection{Overview of the gap distributions} 
We analyze the relative checkpoint-selection gap with respect to the best-observed test performance along the complete training trajectory. A gap equal to zero indicates that the validation-based procedure selected a checkpoint attaining the best-observed test value of the target metric, whereas larger values indicate a greater relative performance loss. The best-observed test value is used exclusively as a retrospective diagnostic reference and never participates in checkpoint selection. The complete empirical gap distributions are reported separately for post-hoc checkpoint selection and patience-based early stopping. 
In this section, Fig.~\ref{fig:uci_ce_f1} provides an illustrative UCI example, comparing post-hoc checkpoint selection and patience-based early stopping under Cross-Entropy training with macro-F1 as the target metric. The plots additionally distinguish the parameter-to-sample ratio $r$. Tables~\ref{tab:selection_gap} and~\ref{tab:selection_gap_es_p95} summarize the upper tail of these distributions through their empirical 95th percentile. The remaining UCI configurations are reported in Appendix~\ref{app:additional_uci}. All conclusions reported below are based on the full set of experimental configurations and on the distribution summaries reported in Tabs. \ref{tab:selection_gap} and \ref{tab:selection_gap_es_p95}.
For the image datasets, Figs. ~\ref{fig:image_posthoc} and ~\ref{fig:image_earlystopping} report the complete image-classification results for post-hoc checkpoint selection and patience-based early stopping, respectively, with accuracy, macro-F1, and MCC shown in separate panels.

\subsection{Post-hoc checkpoint selection} 
\label{subsec:results_posthoc}  

Under post-hoc checkpoint selection, training proceeds for the complete predefined set of epochs $\epochset$ and each validation criterion $\evalfun\in\evalfunset$ is used only to rank the available checkpoints $\theta_e,\ \forall e \in\epochset$. This setting isolates the checkpoint-ranking effect of the validation criterion from the additional effect of terminating training.  

\paragraph{Tabular datasets.} 

The UCI results show that selecting checkpoints directly according to the target validation metric is generally effective. For macro-F1 and MCC, the target metric produces the smallest 95th-percentile gap under each of the considered training losses (see Tab. \ref{tab:selection_gap} and Fig.~\ref{fig:uci_ce_f1}a). For accuracy, it produces the smallest percentile under Cross-Entropy and PolyLoss training, while validation C-Loss performs slightly better under C-Loss training. The advantage of target-metric-based selection is particularly evident for macro-F1. In this case, the corresponding gap distributions are generally more concentrated near zero than those produced by loss-based criteria. This difference is especially pronounced under C-Loss training, for which Cross-Entropy and PolyLoss validation produce broader distributions. A similar, although less marked, pattern is observed for MCC (see Appendix).  Validation C-Loss is the most competitive loss-based criterion in several cases. It occasionally produces the smallest upper-tail gap and, in several other configurations, performs comparably to the target metric. By contrast, matching the training and validation losses does not produce a systematic advantage under post-hoc selection. Alignment is beneficial in some configurations but is outperformed in others by either the target metric or a different validation loss.

\paragraph{Image datasets.}
The image results provide stronger evidence in favour of validation criteria aligned with the target metric (see Tab. \ref{tab:selection_gap} and Fig.~\ref{fig:image_posthoc}). Under C-Loss training, the target metric produces the smallest 95th-percentile gap for accuracy, macro-F1, and MCC. The corresponding distributions are substantially more concentrated near zero than those obtained using Cross-Entropy or PolyLoss as validation criteria.  Validation C-Loss is nevertheless highly competitive when models are trained using Cross-Entropy or PolyLoss. Under these training losses, it frequently produces the smallest gap, particularly for accuracy and macro-F1. For MCC, both validation C-Loss and the target metric yield distributions concentrated close to zero.  A different behaviour arises when image models are trained using C-Loss and checkpoints are selected according to Cross-Entropy or PolyLoss. These combinations produce broad distributions and gaps, particularly for accuracy and macro-F1. The strip plots indicate that this effect is especially evident on TinyImageNet-200 (see Fig.~\ref{fig:image_posthoc}). Overall, the post-hoc results identify the target metric and C-Loss as the most reliable validation criteria in the considered settings. The target metric is more frequently associated with the smallest upper-tail gap, whereas C-Loss provides a competitive alternative in several configurations. Neither criterion, however, dominates across all training losses, target metrics, and datasets.

\subsection{Patience-based early stopping}
Patience-based early stopping assigns the validation criterion a dual role: it determines both when training is terminated and which checkpoint is retained among those observed up to that point. Consequently, the resulting performance gap may arise either because the criterion fails to identify the best checkpoint with the
best target-metric value among those already observed or because training stops before better-performing checkpoints are reached. 

\paragraph{Tabular datasets.}

Under patience-based early stopping, the large checkpoint selection gap observed on several datasets (see Tab. \ref{tab:selection_gap_es_p95} and Fig.~\ref{fig:uci_ce_f1}b) may result from two distinct effects: the selection criterion may choose a suboptimal checkpoint among those already available, or training may stop before better checkpoints are reached. In practice, both effects may also occur simultaneously. Consequently, the post-hoc ranking of validation criteria does not necessarily carry over to the early-stopping setting.
On the UCI datasets, early stopping generally produces broader gap distributions and larger upper tails than post-hoc checkpoint selection. The difference is especially evident when the target metric itself is used as the early stopping signal. Although target-metric-based selection performs well when the complete trajectory is available, its performance becomes less stable when the same metric also determines the training stop. For macro-F1 and MCC, matching the validation loss used for the checkpoint selection to the training loss produces the smallest 95th-percentile gap under each of the three training objectives. This pattern contrasts with the post-hoc results, where loss alignment does not provide a consistent advantage. Accuracy exhibits a different pattern. Validation C-Loss produces the smallest upper-tail gap under each training loss, although the differences among criteria are limited in some configurations. Summarizing, loss alignment is most effective for macro-F1 and MCC, whereas validation C-Loss is more reliable for accuracy. These findings show that a criterion that performs well under post-hoc selection may still terminate training prematurely or retain a less effective checkpoint when used within a patience-based protocol.

\paragraph{Image datasets.}

For image classification, under Cross-Entropy and PolyLoss training, validation C-Loss consistently produces the smallest 95th-percentile gap across the considered target metrics (see Tab.~\ref{tab:selection_gap_es_p95} and Fig.~\ref{fig:image_earlystopping}). Target-metric-based early stopping also remains competitive in these settings, and the corresponding distributions are generally concentrated near zero.  A substantially less stable behaviour is observed under C-Loss training. For accuracy and MCC, monitoring the corresponding target metric produces the smallest  gap. Cross-Entropy and PolyLoss instead lead to broader distributions, indicating that these validation signals may terminate C-Loss training before high-performing checkpoints are reached.  Macro-F1 constitutes the most critical configuration. Under C-Loss training, all validation criteria produce substantially larger gaps than in the corresponding post-hoc setting. In this setting, target-metric-based early stopping is not the best choice in several cases, providing a direct counterexample to the claim that monitoring the target metric is always preferable.  Across the image experiments, validation C-Loss is the most frequently effective early stopping criterion, but its advantage is not universal. Under C-Loss training, the preferred criterion depends strongly on the target metric, and even the best-performing criterion may exhibit a substantial upper tail.

\subsection{Dataset effects}
\label{subsec:dataset_parameterization_results}  
The complete distributions reveal considerable heterogeneity across datasets. Under post-hoc selection, several datasets exhibit gaps concentrated near zero across most validation criteria and parameter-to-sample ratios. Other datasets show broader distributions and greater sensitivity to the validation criterion. This heterogeneity becomes more pronounced under early stopping, particularly under C-Loss training. Ordering the datasets according to GDV does not reveal a uniform monotonic relationship between linear separability and the checkpoint-selection gap. Datasets occupying similar positions in the GDV ordering can exhibit markedly different distributions, while the effect of changing the validation criterion varies across datasets. The present experimental assessment does not support treating GDV as an independent predictor of checkpoint-selection reliability. The parameter-to-sample ratio $r$ also interacts with the dataset, training loss, target metric, and selection protocol. However, for some datasets, the point clouds corresponding to different values of $r$ are visibly separated (see Fig.~\ref{fig:uci_ce_f1}), indicating that the ratio between number of parameters and samples materially affects checkpoint selection. For others, the distributions largely overlap. Moreover, the direction of the effect is not consistent across experimental configurations: increasing $r$ may be related to a reduction of the gap, or to an increase, or have little observable effect. Consequently, the results do not identify a regime of $r$ that is uniformly preferable for checkpoint selection.

\subsection{Discussion} 
\label{subsec:discussion}  
The results do not establish a universal ranking of validation criteria. Under post-hoc checkpoint selection, direct selection according to the target validation metric is generally effective and most frequently produces the smallest upper-tail gap. Validation C-Loss is nevertheless competitive and occasionally provides better checkpoint selection.  Under patience-based early stopping, the ranking changes substantially. Loss-based criteria, and especially validation C-Loss, more frequently control the upper tail of the gap distribution, whereas monitoring the target metric can be unstable in some configurations. Matching the training and validation losses also becomes more relevant, particularly for macro-F1 and MCC on tabular datasets.  The distinction between post-hoc selection and early stopping is therefore empirically consequential. A criterion may rank the checkpoints of a completed trajectory effectively while being less suitable for determining when that trajectory should terminate. Conversely, a criterion that is not the best post-hoc ranking signal may provide more reliable early stopping behaviour. The validation criterion should consequently be chosen jointly with the training loss, target metric, dataset, and stopping protocol.

\subsection*{Limitations}
Although the experiments cover both tabular and image-classification settings, the scope of the study remains necessarily limited. The experiments on tabular data employ shallow neural networks to permit controlled variation of the parameter-to-sample ratio, whereas the image experiments use a single deep architecture, namely ResNet50. Consequently, the reported findings should not be assumed to generalize to arbitrary architectures or substantially larger models and datasets.

The size of the validation set may also affect the variance and reliability of both loss-based and metric-based checkpoint evaluation, therefore the tabular dataset experiments exclude datasets with fewer than 1,000 instances. However, in this work the validation-set size is not varied independently of the dataset and architecture. The present results therefore do not isolate its specific contribution to the observed checkpoint-selection gaps.

Moreover, the optimization setup is kept fixed to isolate the effects of the training loss, validation criterion, target metric, model parameterization, and selection protocol. Different optimizers, learning-rate schedules, regularization strategies, or patience values may produce different checkpoint sequences and therefore alter the relative behaviour of the validation criteria. Finally, the analysis considers accuracy, macro-F1, and MCC as target metrics. Although these metrics cover different aspects of classification performance, the study does not address other common metrics such as AUROC or expected calibration error. 

The present experiments do not include transformer-based architectures, natural-language processing tasks, or large-scale tabular applications. The reported findings should therefore not be assumed to transfer directly to these models and domains. Moreover, the study is intentionally restricted to commonly adopted individual validation criteria and does not evaluate combined loss--metric criteria, calibration-aware criteria, multi-objective checkpoint selection, or learned and meta-learning-based early-stopping strategies. These approaches fall outside the scope of the present empirical comparison but constitute important directions for future work. More generally, future studies should assess the robustness of the observed interactions across a broader range of architectures, optimization settings, datasets, validation-set sizes, stopping configurations, and evaluation objectives.
\end{insertedtext}


\ucipairedplot
{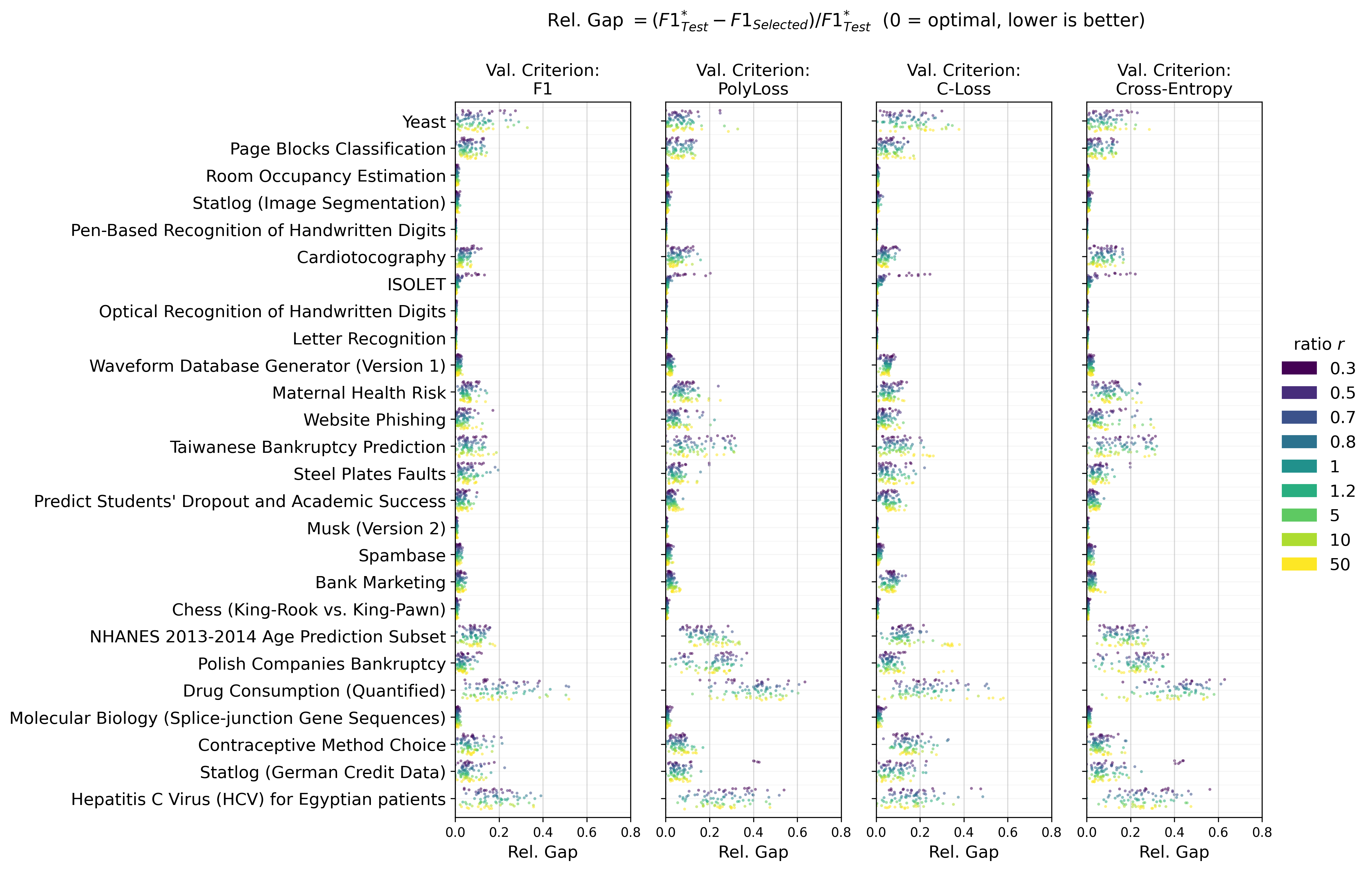}
{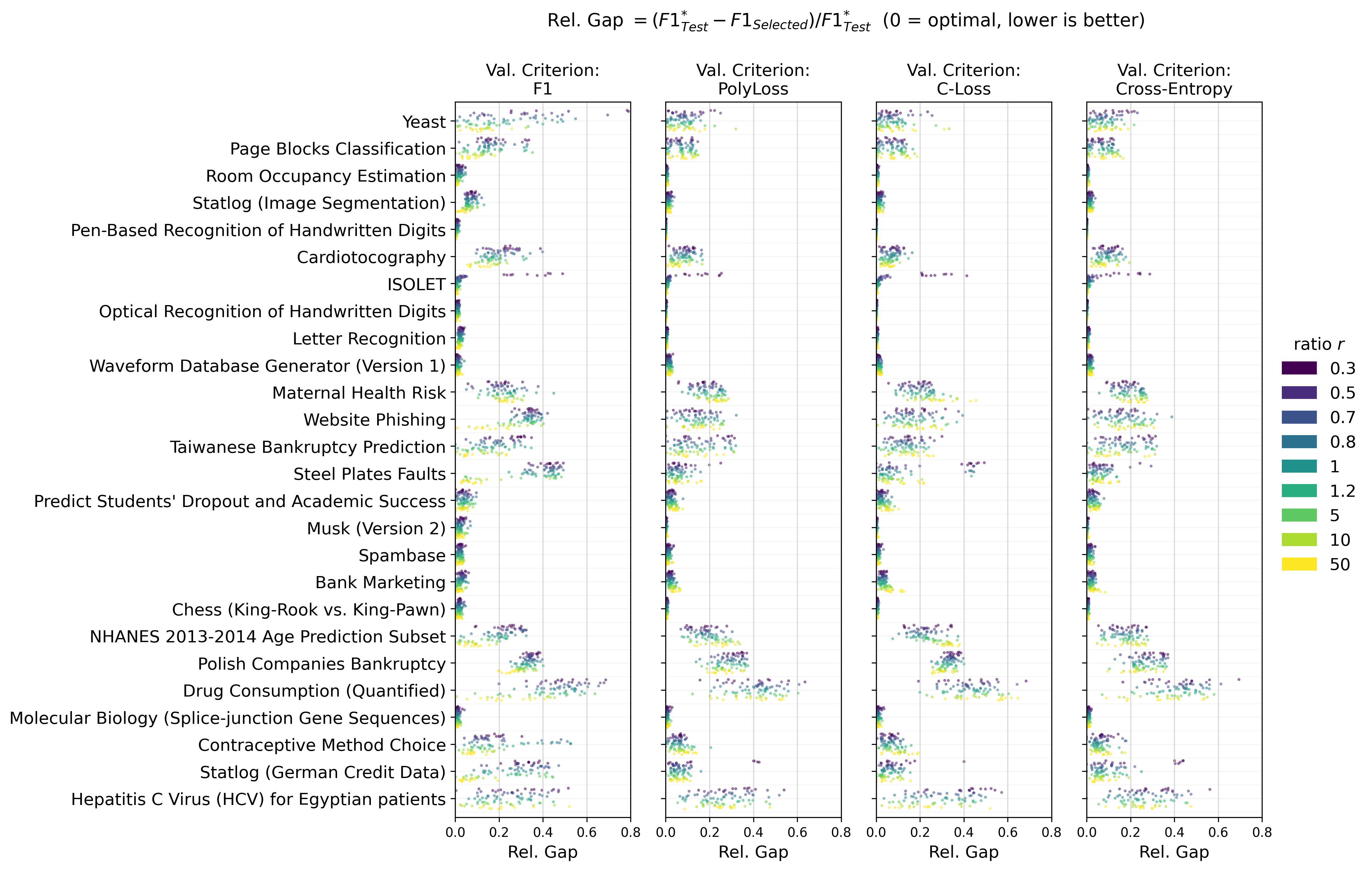}
{Cross-Entropy}
{macro-F1}
{fig:uci_ce_f1}


\section{Conclusions}
\label{sec:conclusions}
\begin{insertedtext}
This study investigated how the loss used to train a classifier, the validation criterion used to select checkpoints, the target evaluation metric, and the stopping protocol jointly affect the performance of the retained model. Rather than asking only whether different validation criteria select different checkpoints, we quantified the resulting target-metric performance gap with respect to the best-observed test performance along a training trajectory. Under post-hoc checkpoint selection, directly ranking checkpoints on a validation set according to the target metric is generally effective and most frequently produces the smallest gap. Validation C-Loss is nevertheless competitive and, in several configurations, performs comparably to or better than the target metric. By contrast, matching the training and validation losses does not provide a systematic advantage in a post-hoc selection protocol. The ranking changes substantially under patience-based early stopping. In particular, using C-Loss as the validation criterion is frequently associated with smaller gaps under early stopping. Alignment between the training and validation losses also becomes more relevant in some tabular configurations. However, these patterns remain conditional on the target metric and training loss. The image experiments under C-Loss training provide a particularly clear example of this dependence, with markedly different behaviour across accuracy, macro-F1, and MCC. Summarizing, under post-hoc checkpoint selection, using the target metric as the validation criterion yields the best results in most configurations. Under early stopping, matching the validation loss to the training loss generally becomes more advantageous, especially for macro-F1 and MCC on the tabular datasets. This pattern is not universal, however: validation C-Loss is often preferable for accuracy and image classification, while the target metric remains the best criterion in some configurations involving C-Loss training. The experiments also reveal that neither dataset separability nor the parameter-to-sample ratio alone explains the observed checkpoint-selection gaps. Their effects depend on the training loss, validation criterion, target metric, and stopping protocol. From a practical perspective, the validation criterion should not be chosen independently of its operational role. A criterion used to rank in a post-hoc selection protocol need not be the most appropriate signal for deciding when training should terminate. Similarly, using the target metric as the validation criterion does not guarantee reliable early stopping, despite its direct alignment with the final evaluation objective. Training loss, target metric, and stopping protocol should therefore be considered jointly when designing a validation-based model selection procedure. The present analysis remains empirical and is limited to the considered classification datasets, architectures, training losses, target metrics, and patience-based stopping configuration. Moreover, the best-observed test performance is defined relative to a number of training epochs given a priori and should not be interpreted as a globally optimal or deployable model selection target. Future work should investigate whether the observed interactions persist across additional architectures, validation-set sizes, optimization procedures, and application domains.  It would also be valuable to study stopping rules that use separate functions for trajectory termination and checkpoint selection, together with combined loss--metric criteria, calibration-aware and multi-objective checkpoint selection, and learned validation or stopping strategies. Overall, the main conclusion is not that one validation criterion should always replace another.
Rather, checkpoint selection and early stopping constitute related but distinct decisions, and the effectiveness of a validation criterion depends critically on which of these decisions it is intended to support. Moreover, these results highlight the need for further research into validation criteria that provide a more reliable assessment of a model's generalization ability, regardless of whether they are used for post-hoc checkpoint selection, early stopping, or both.
\end{insertedtext}

\section*{Acknowledgment}
This work was partially funded by the PNRR MUR project PE0000013-FAIR (CUP: E63C25000630006).

\newpage
\bibliography{__BIB}
\begin{appendices}
\newpage
\appendix
\section{Additional UCI checkpoint-selection gap distributions}
\label{app:additional_uci}

This appendix complements the illustrative UCI configuration reported in Figure~\ref{fig:uci_ce_f1} by presenting the pointwise relative checkpoint-selection gap distributions for all the remaining combinations of training loss and target metric.

\ucipairedplot
{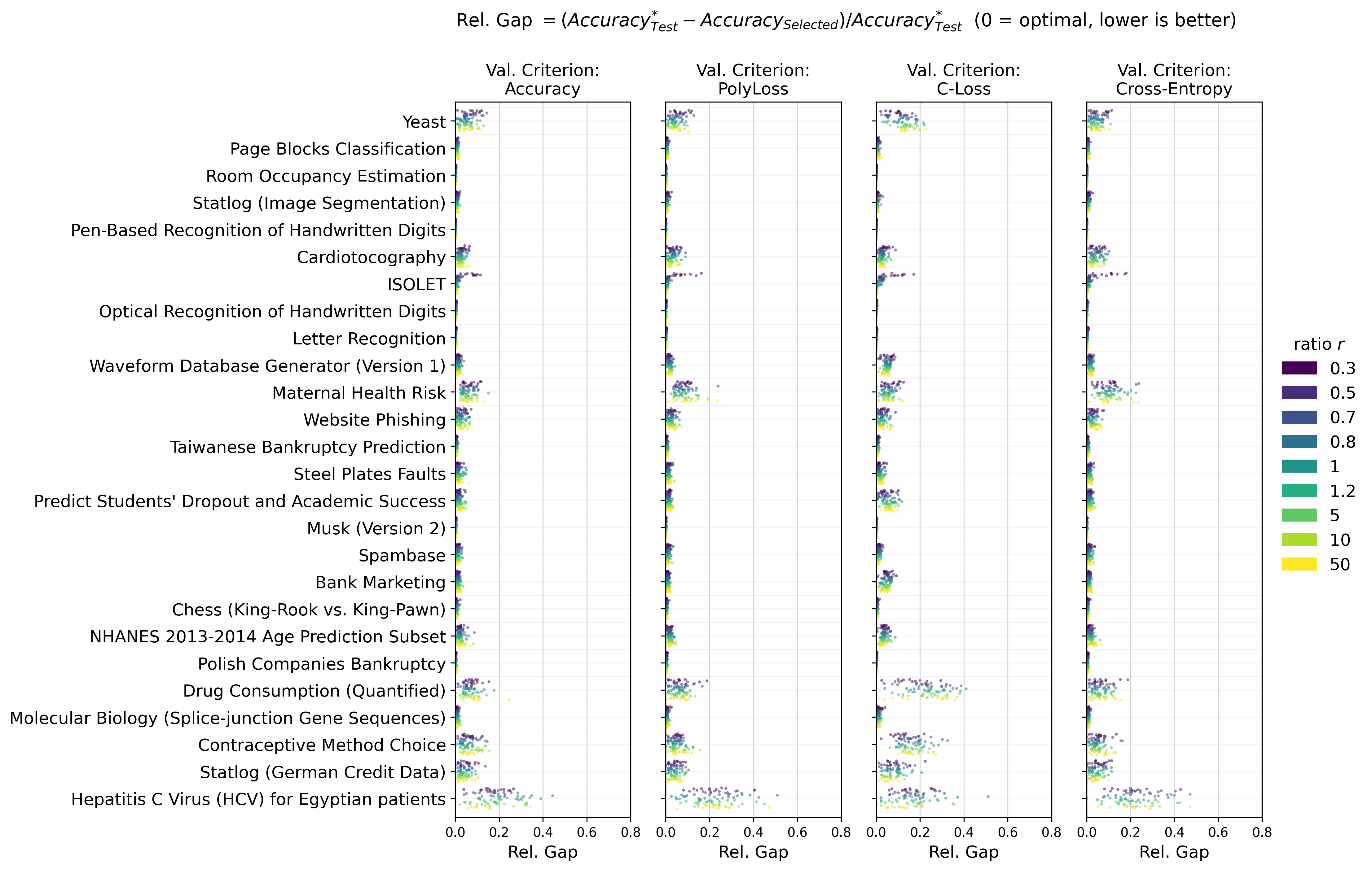}
{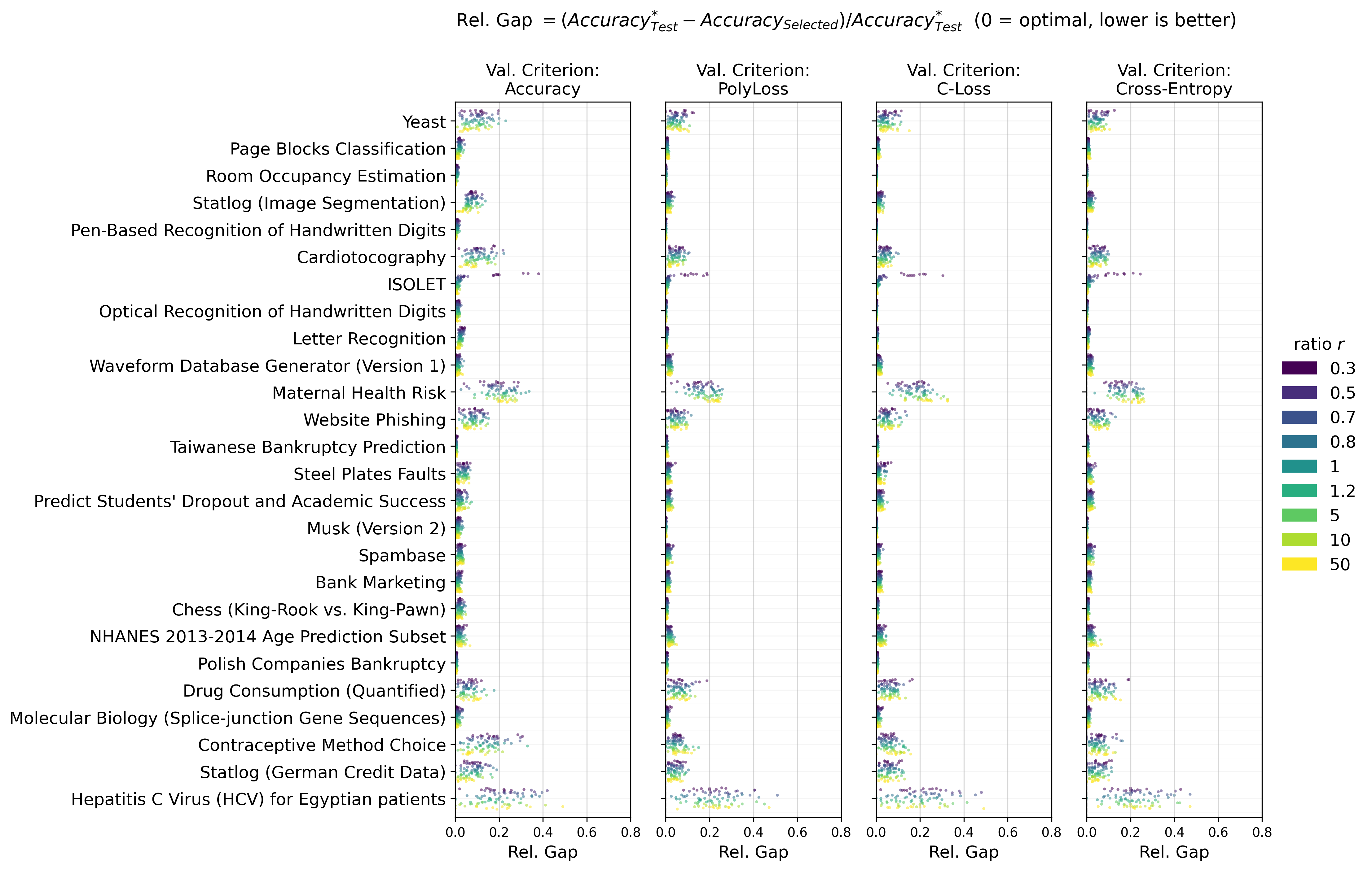}
{Cross-Entropy}
{accuracy}
{fig:uci_ce_acc}

\ucipairedplot
{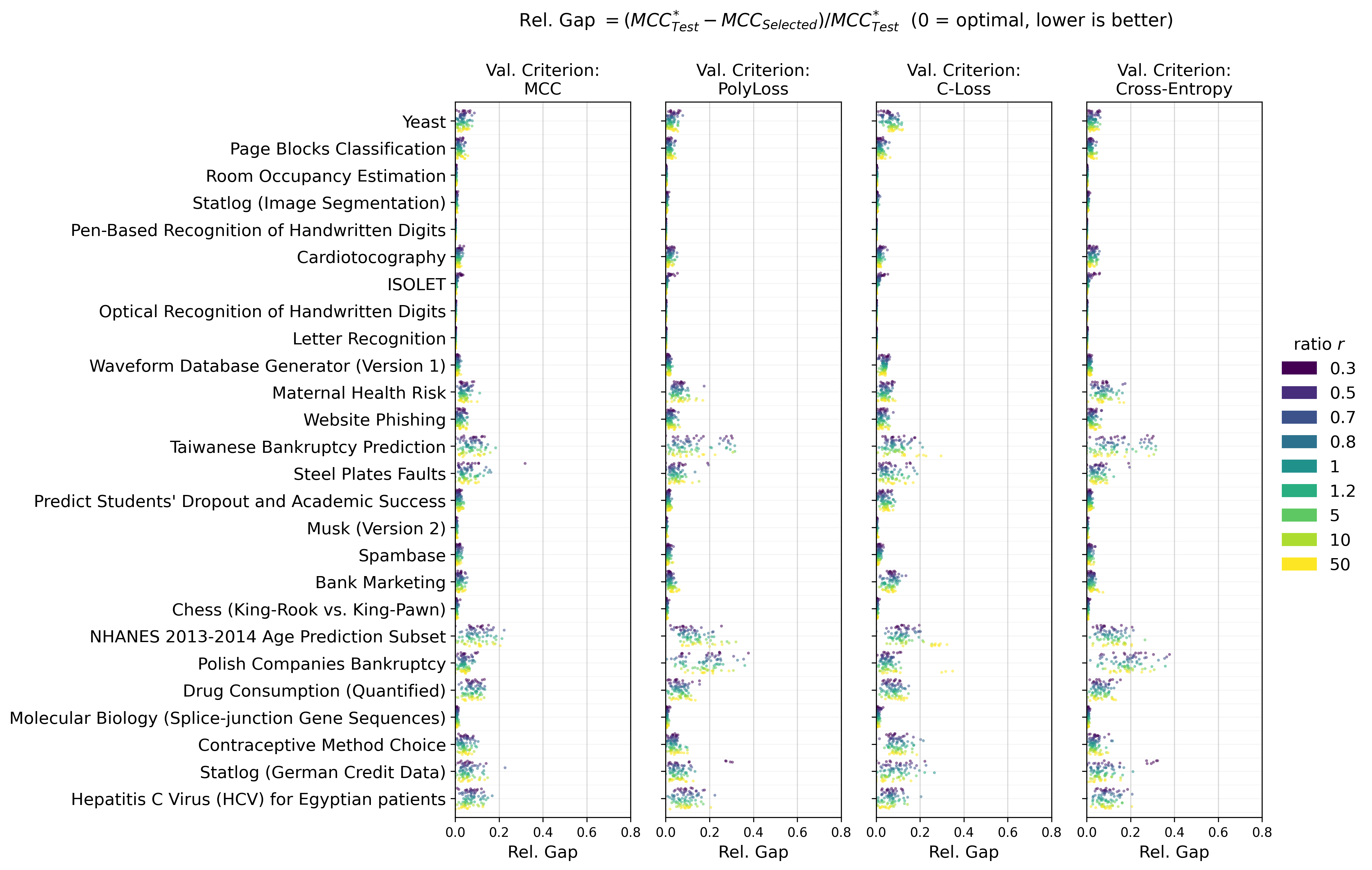}
{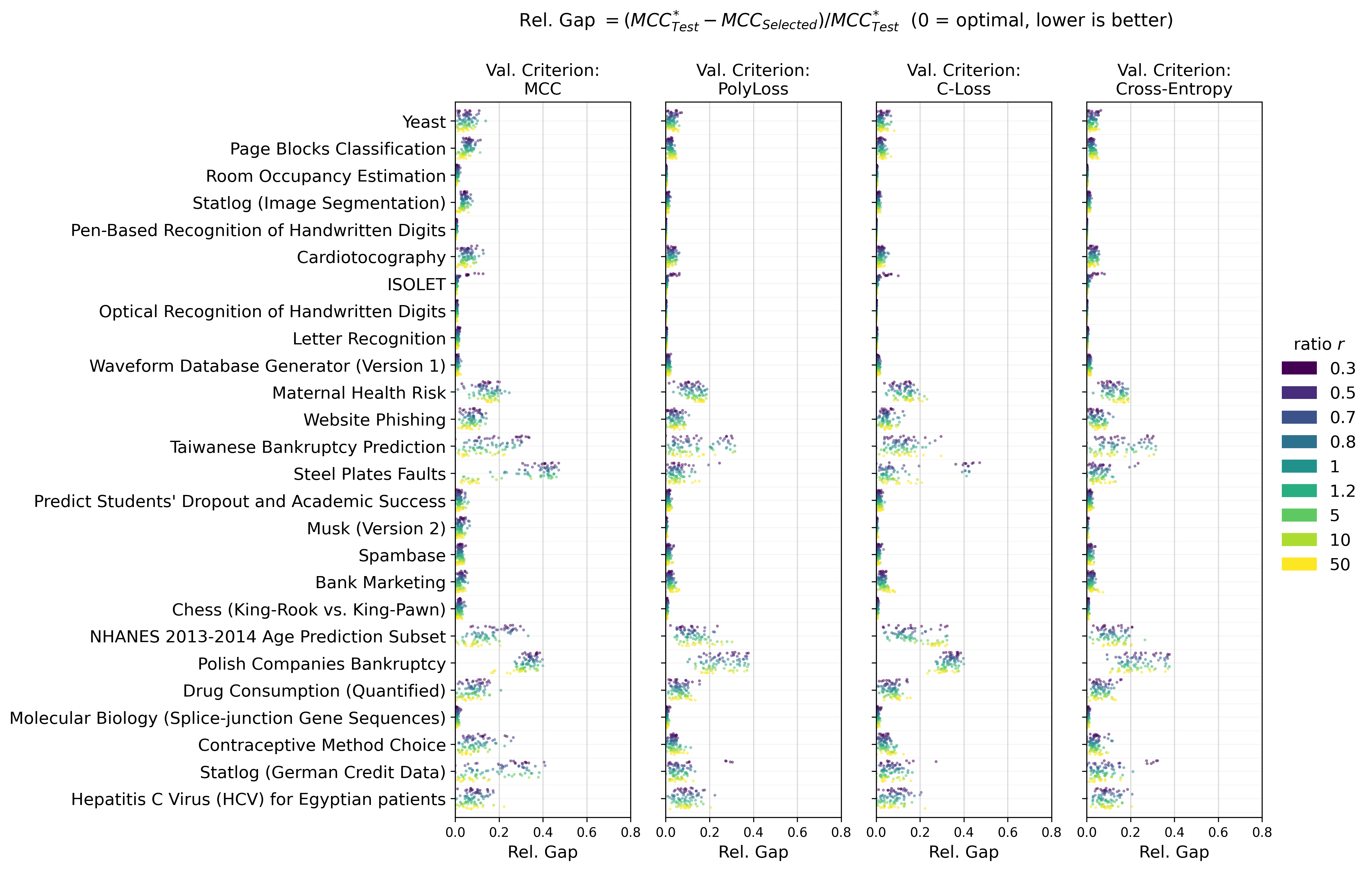}
{Cross-Entropy}
{MCC}
{fig:uci_ce_mcc}

\ucipairedplot
{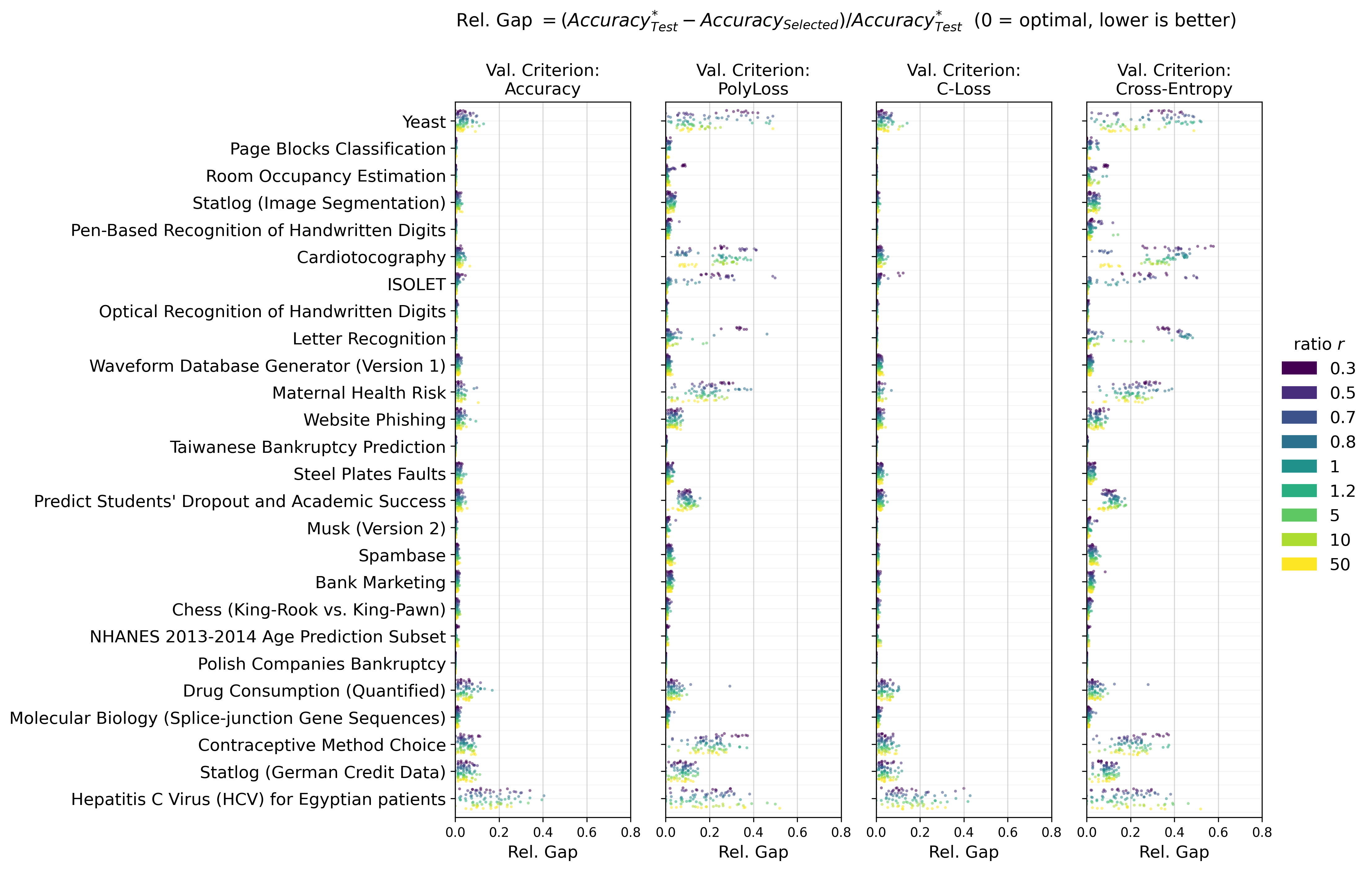}
{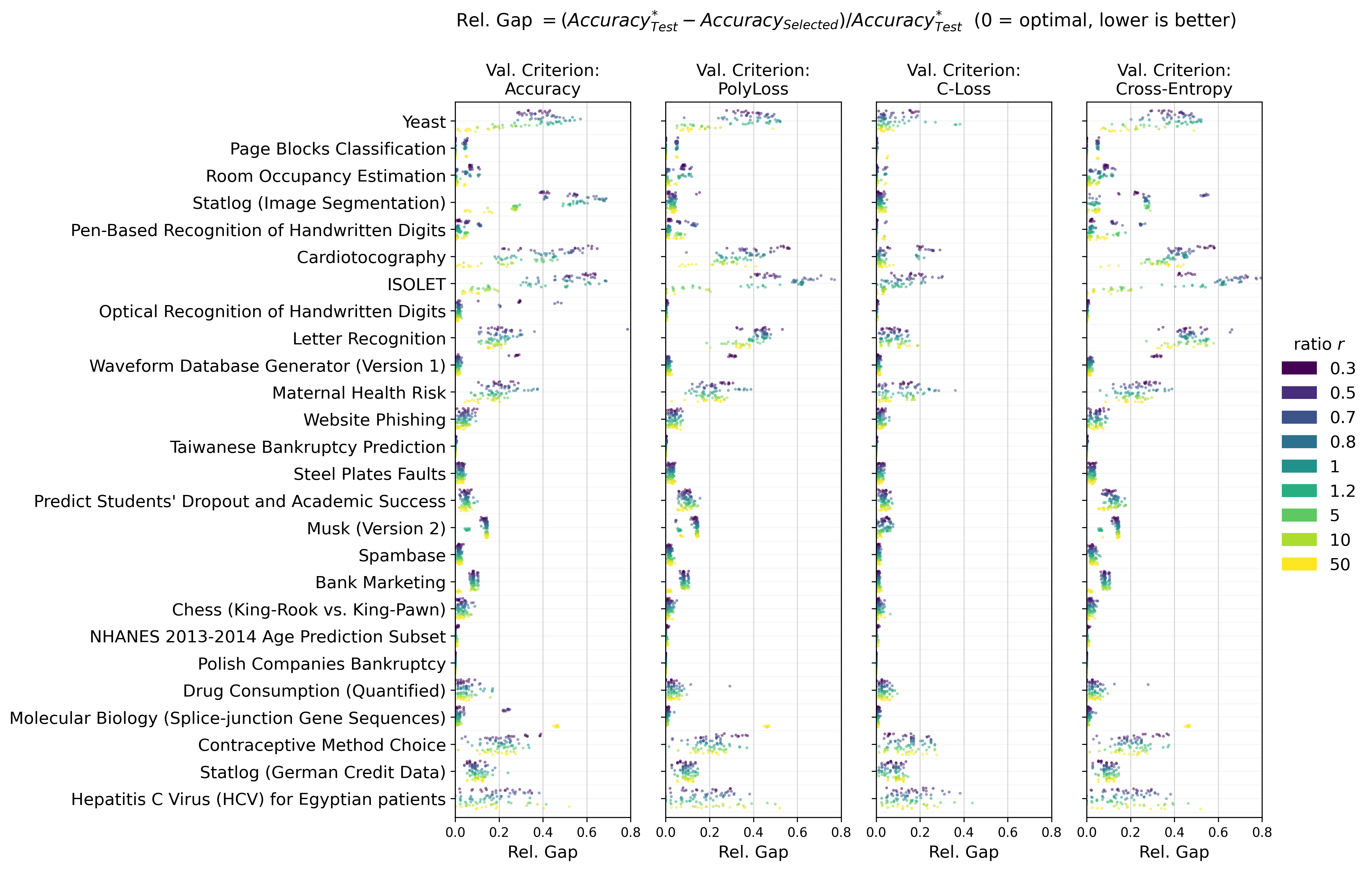}
{C-Loss}
{accuracy}
{fig:uci_closs_acc}

\ucipairedplot
{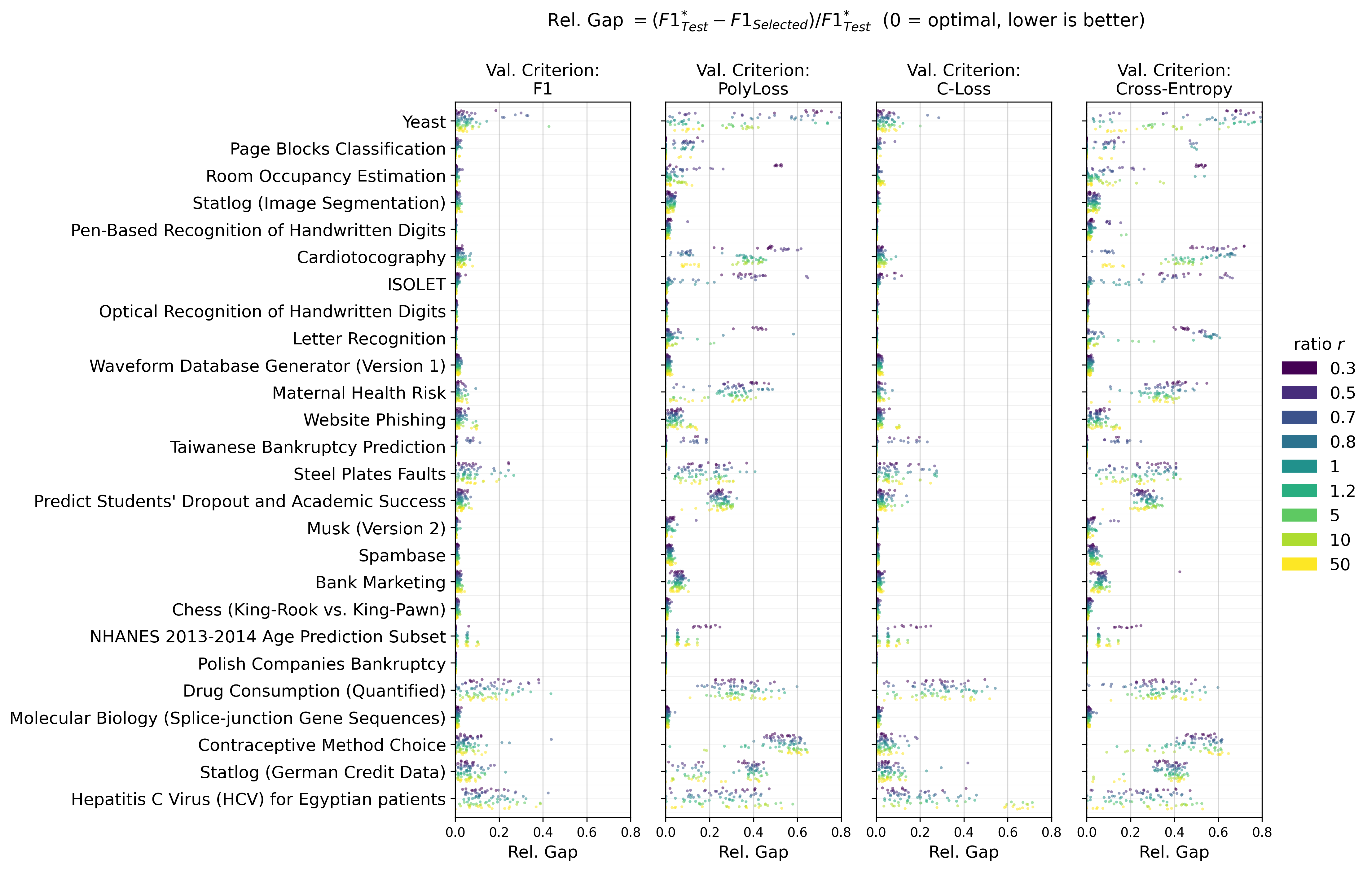}
{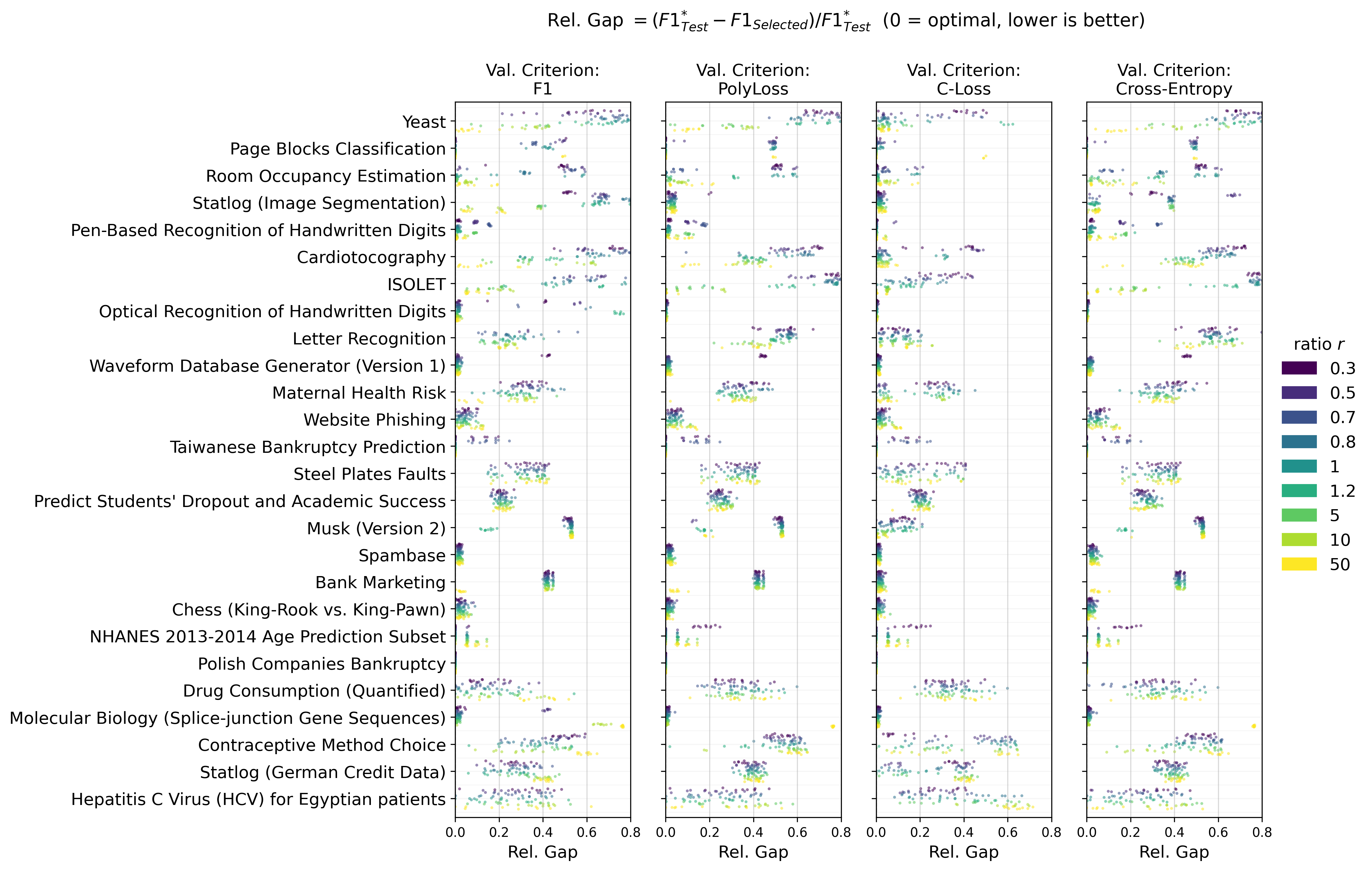}
{C-Loss}
{macro-F1}
{fig:uci_closs_f1}

\ucipairedplot
{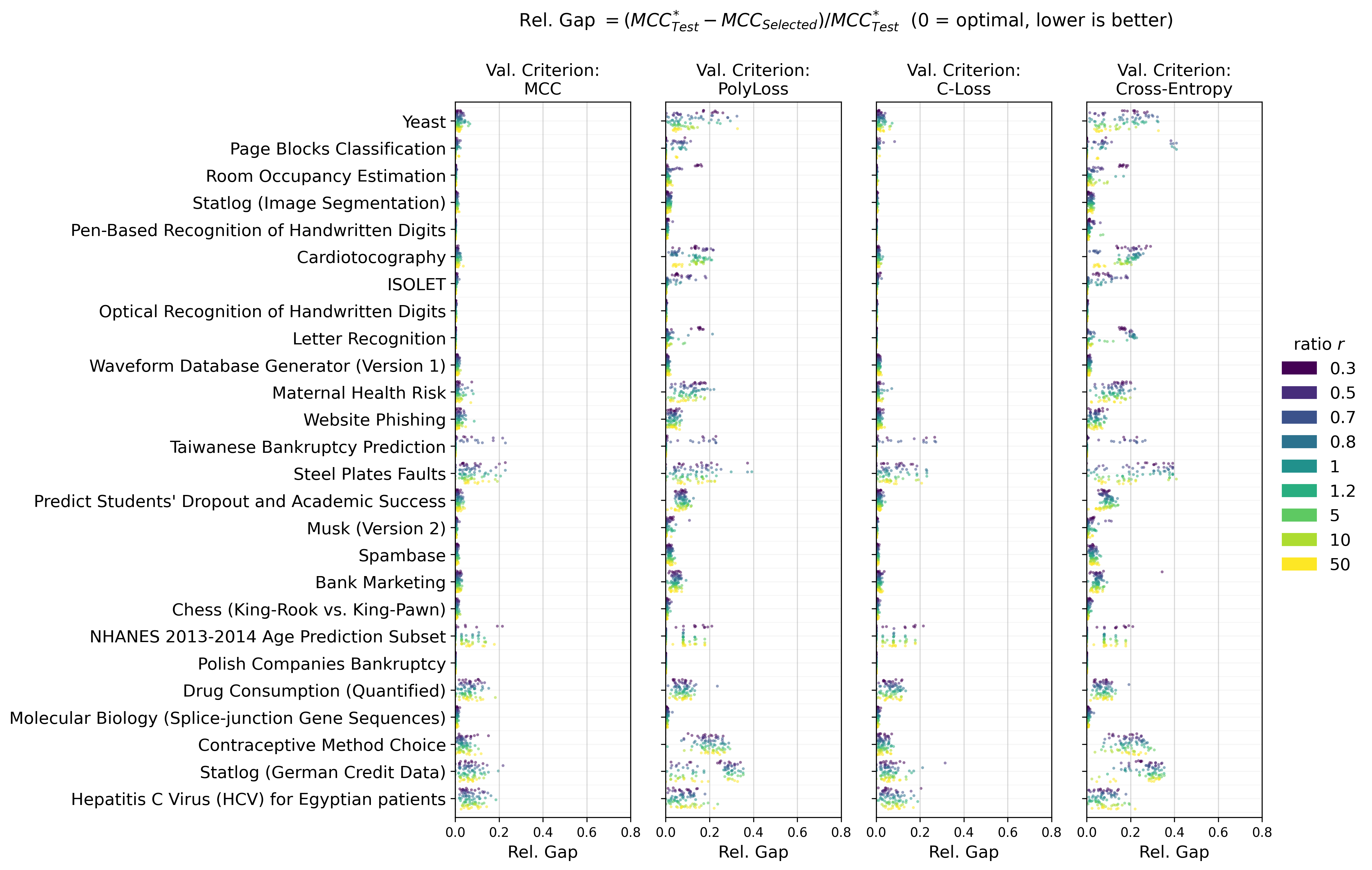}
{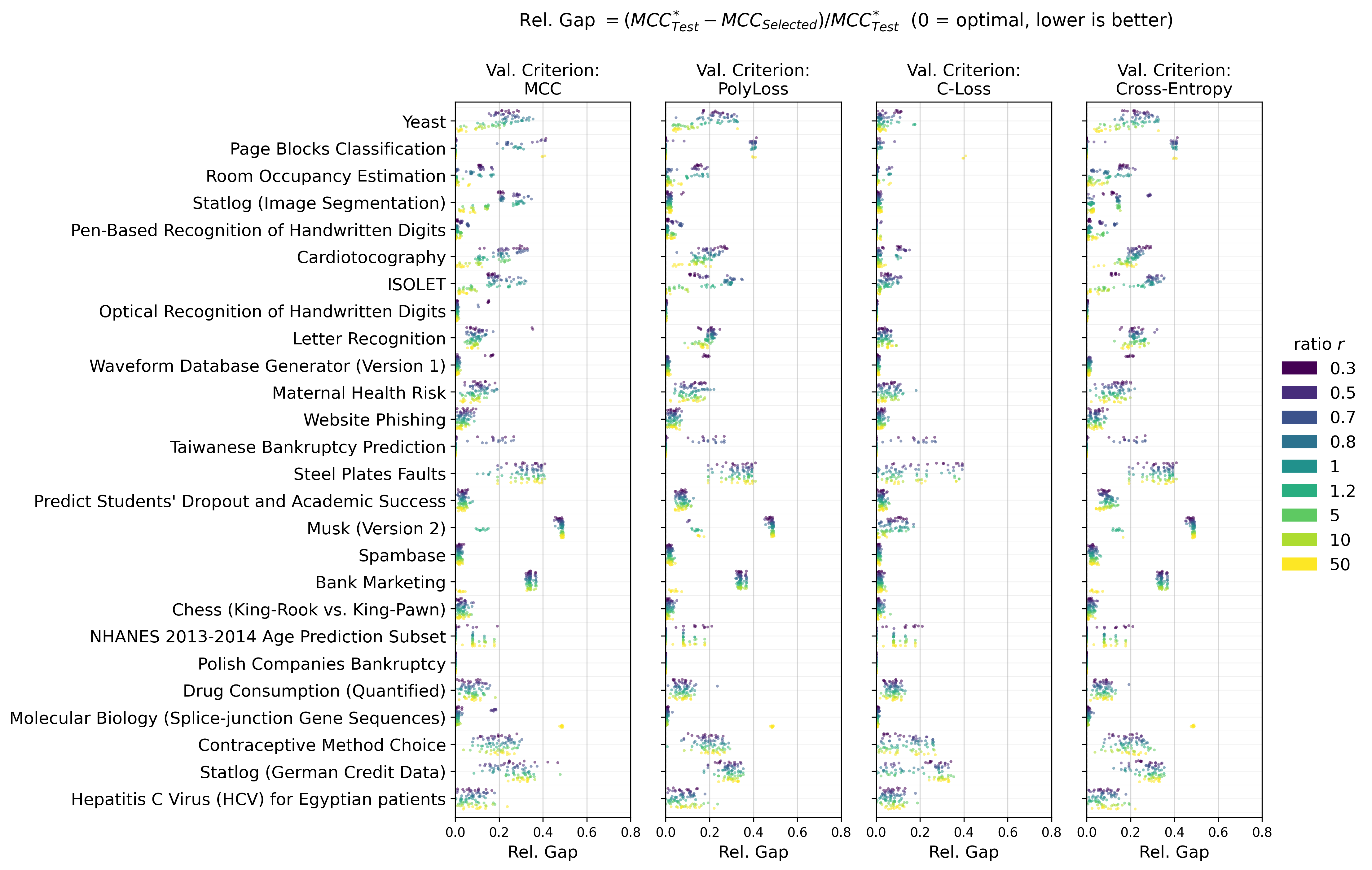}
{C-Loss}
{MCC}
{fig:uci_closs_mcc}

\ucipairedplot
{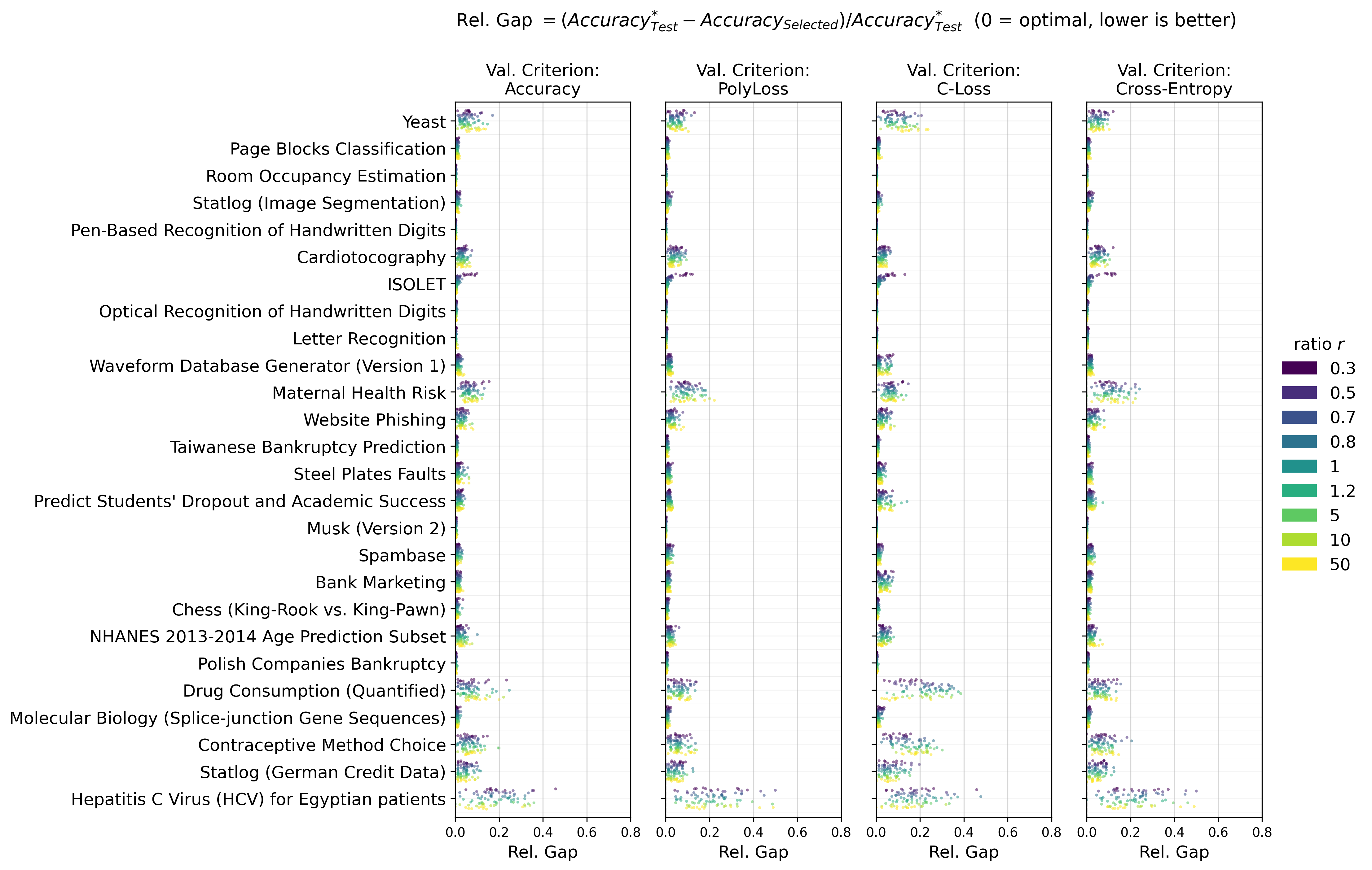}
{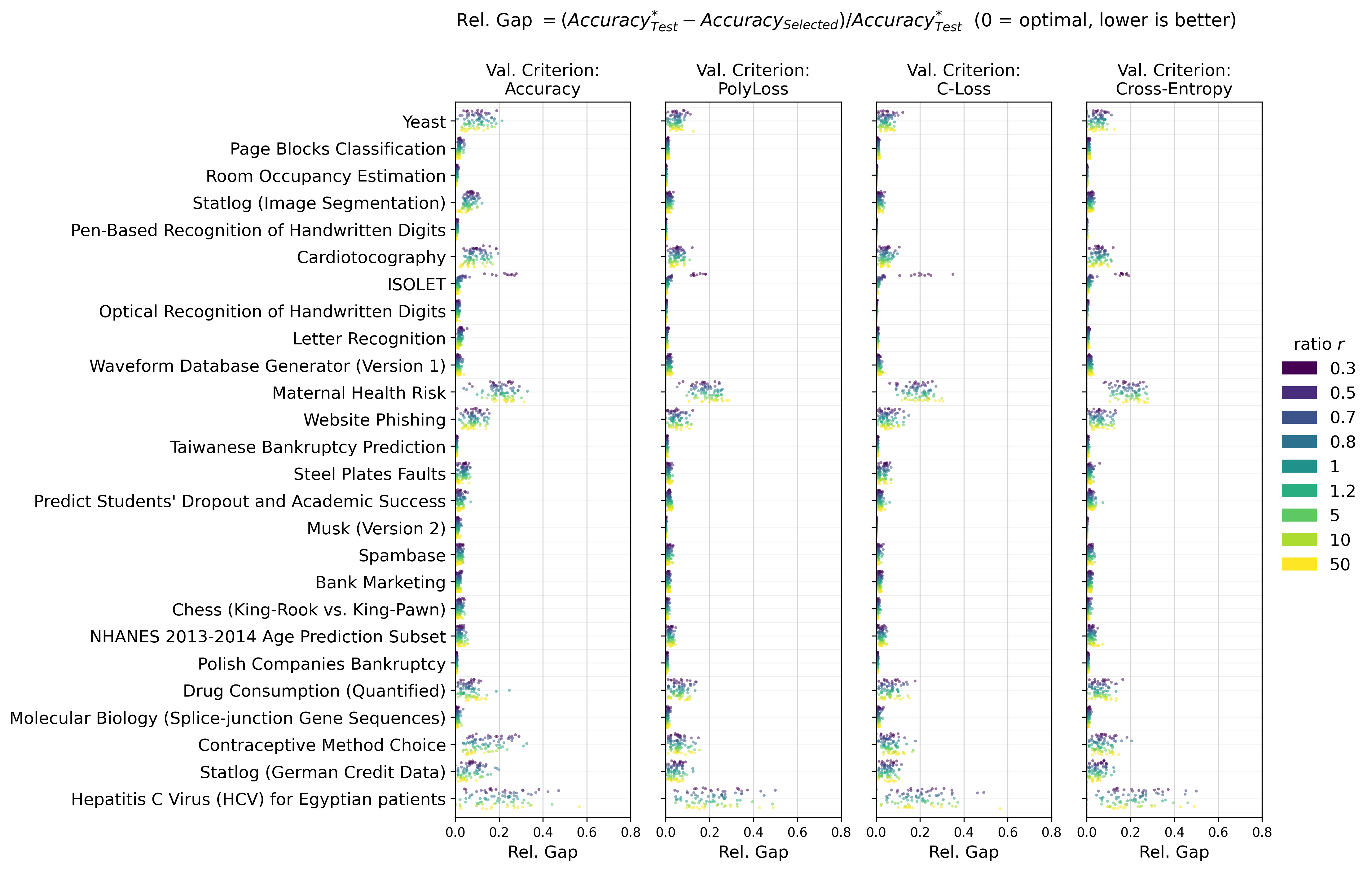}
{PolyLoss}
{accuracy}
{fig:uci_polyloss_acc}

\ucipairedplot
{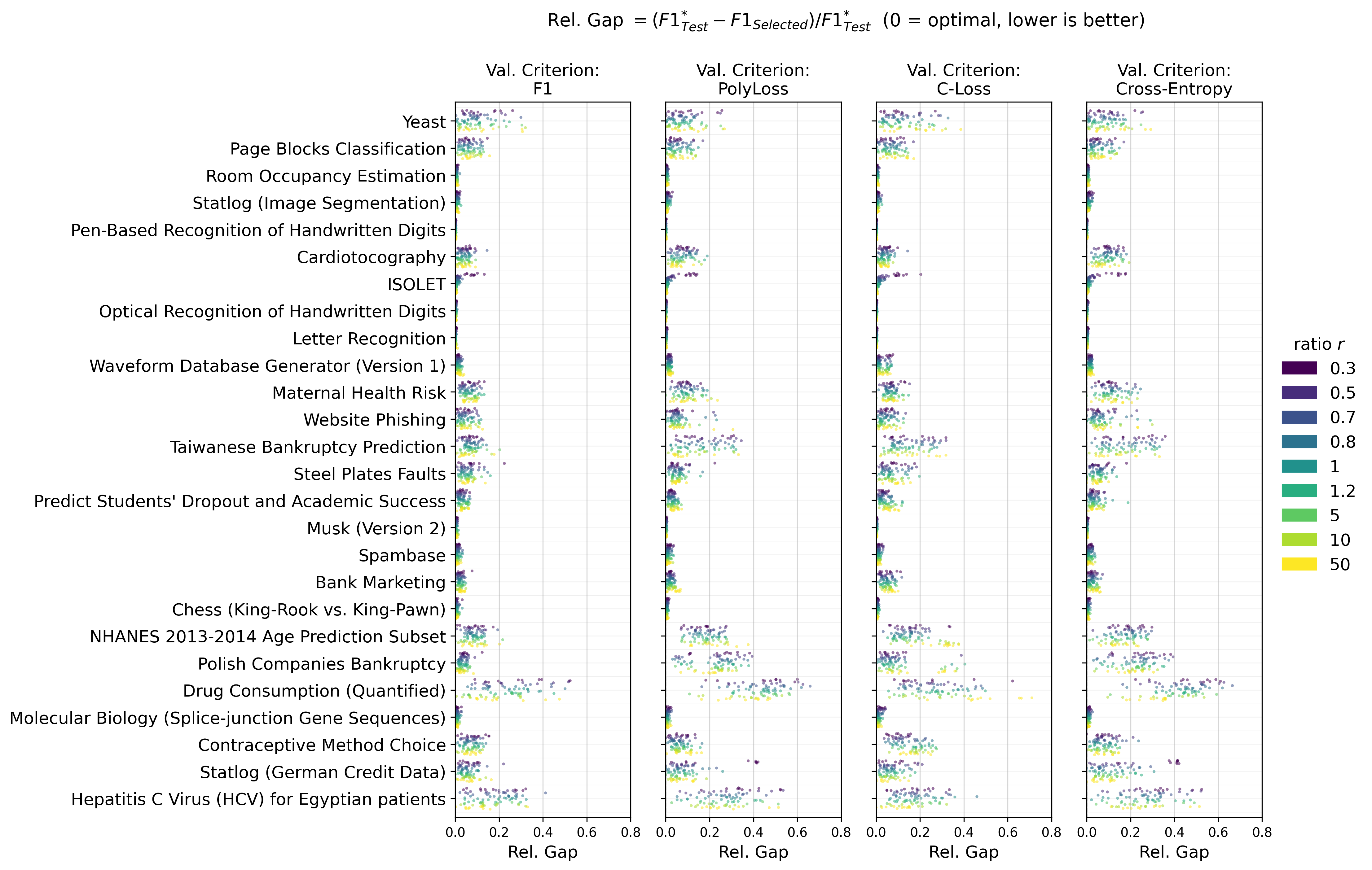}
{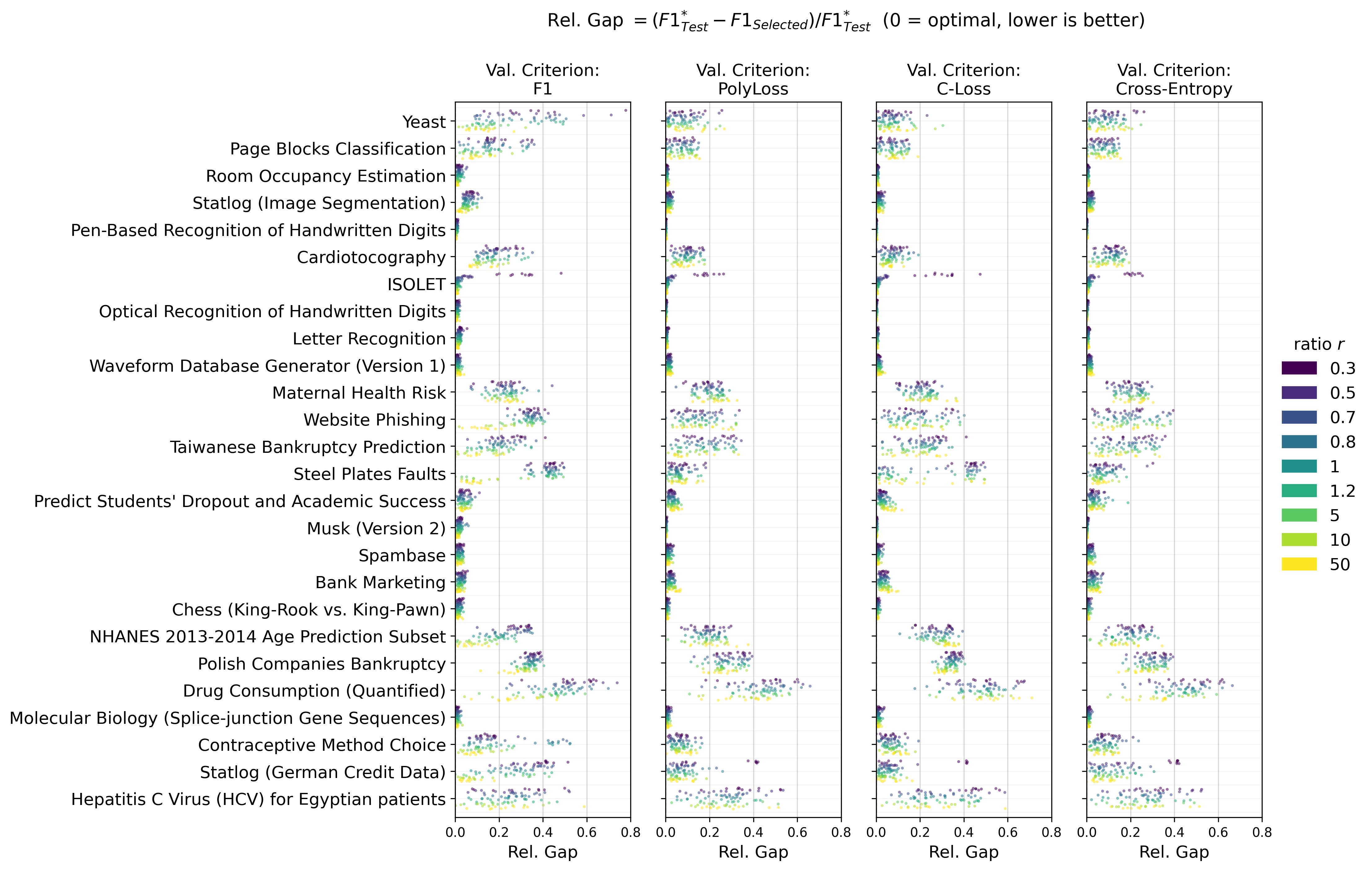}
{PolyLoss}
{macro-F1}
{fig:uci_polyloss_f1}

\ucipairedplot
{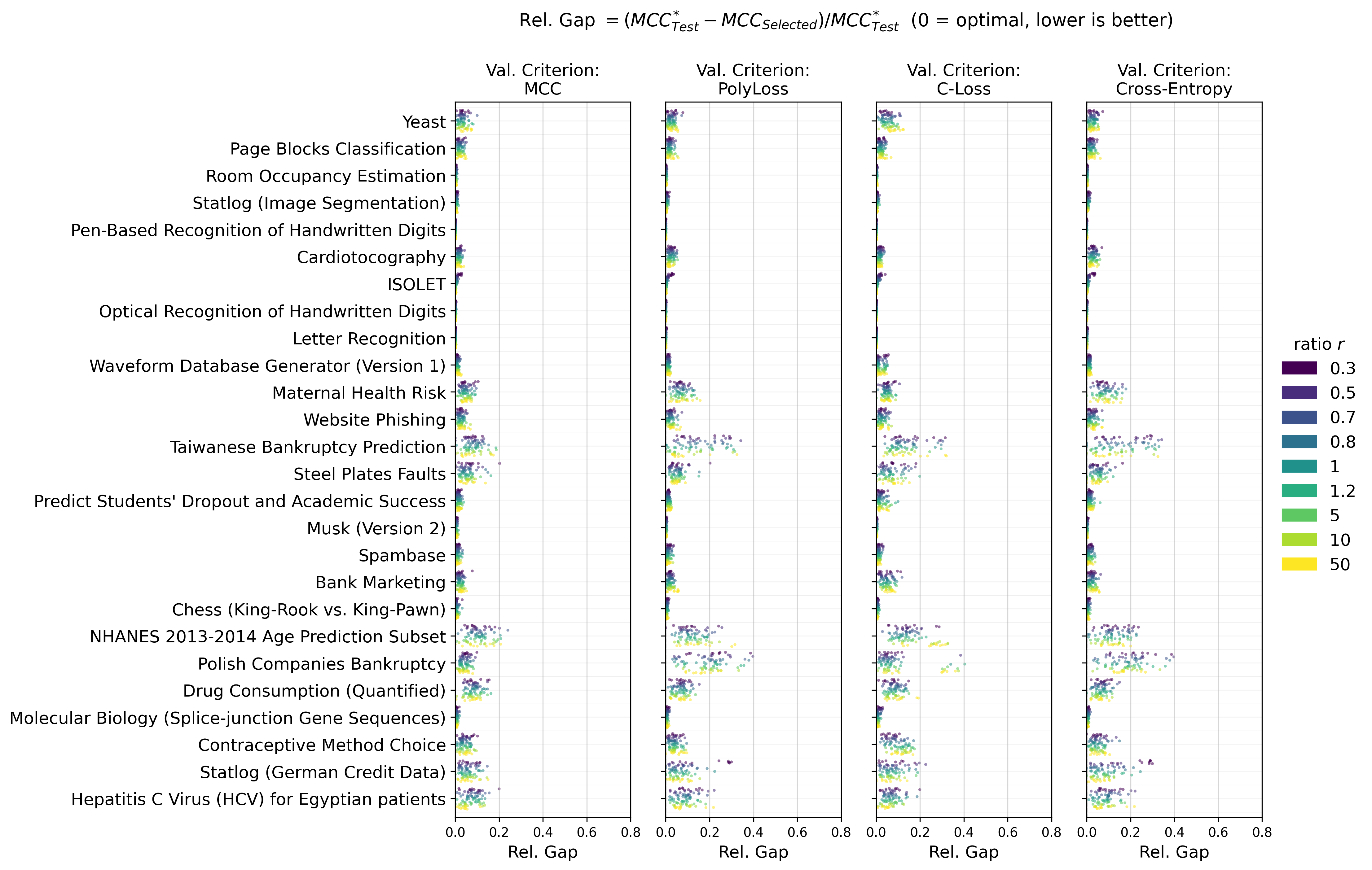}
{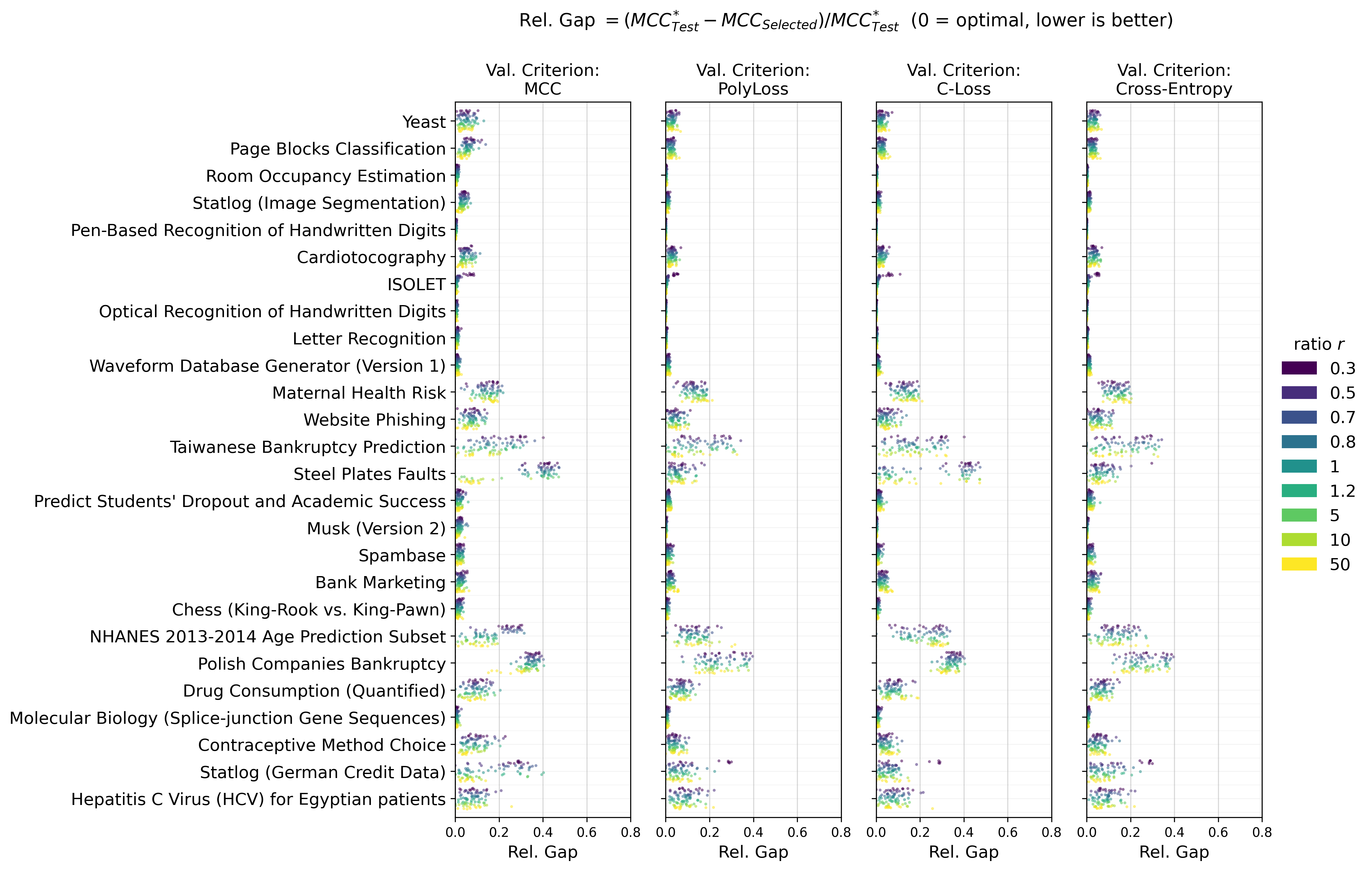}
{PolyLoss}
{MCC}
{fig:uci_polyloss_mcc}

\end{appendices}

\end{document}